\pdfoutput=1

\documentclass[11pt,a4paper]{article}
\usepackage{acl}

\usepackage[most]{tcolorbox}
\tcbuselibrary{skins,breakable}
\usepackage{wrapfig}

\usepackage{latexsym}
\usepackage{graphicx}
\usepackage{booktabs}
\usepackage[T1]{fontenc}

\usepackage[utf8]{inputenc}

\usepackage{microtype}
\usepackage{cleveref}
\usepackage{inconsolata}

\usepackage{graphicx}
\usepackage{fancyhdr}
\usepackage{dblfloatfix} 
\usepackage{pifont}      
\usepackage{array}       
\usepackage{makecell}    
\usepackage{ragged2e}    
\usepackage{rotating}
\usepackage{tabularx}
\usepackage{colortbl}
\usepackage{multirow}
\usepackage{enumerate}
\usepackage{longtable}
\usepackage{url}         
\usepackage{seqsplit}
\usepackage{pdflscape}   
\usepackage{graphicx}
\usepackage{subcaption}
\usepackage{threeparttable} 
\usepackage{placeins}    

\usepackage{caption}     
\usepackage{cuted} 

\usepackage[sfdefault]{FiraSans}

\newcolumntype{P}[1]{>{\centering\arraybackslash}p{#1}}
\newcolumntype{M}[1]{>{\centering\arraybackslash}m{#1}}
\newcolumntype{L}[1]{>{\raggedright\arraybackslash}m{#1}}
\newcolumntype{C}[1]{>{\centering\arraybackslash}m{#1}}

\usepackage{xcolor}
\definecolor{mypurple}{RGB}{112, 114, 234}
\usepackage{enumitem}
\setlist[description]{leftmargin=1.2em,labelsep=0.4em,itemsep=2pt,topsep=2pt,font=\bfseries}

\usepackage{xcolor}
\definecolor{darkgreen}{rgb}{0.0,0.4,0.0}

\usepackage{xspace}
\newcommand{\apex}{APEX\textendash Accounting\xspace}

\newcommand{\NModels}{9}
\newcommand{\NTasks}{160}
\newcommand{\NWorlds}{10}
\newcommand{\TasksPerWorld}{16}
\newcommand{\DevSetTasks}{10}
\newcommand{\NRuns}{8}              
\newcommand{\KSubsample}{3}         
\newcommand{\NTrajectories}{11{,}520} 

\newcommand{\TopModel}{Claude-Fable-5}
\newcommand{\TopScore}{$56.4\%$}              

\newcommand{\SecondModel}{Claude-Opus-4.8}
\newcommand{\SecondScore}{$48.0\%$}
\newcommand{\ThirdModel}{GPT-5.6-Sol}       
\newcommand{\ThirdScore}{$51.5\%$}
\newcommand{\TopPassEight}{$20.1\%$}          
\newcommand{\MaxPassK}{$2.6\%$}               

\newcommand{\NExperts}{42}
\newcommand{\ExpertMedianYears}{11}
\newcommand{\ExpertMeanYears}{12.6}
\newcommand{\BigFourPct}{52.4\%}
\newcommand{\MeanExpertsPerWorld}{14}

\newcommand{\MeanFilesPerTask}{7.5}
\newcommand{\MeanFilesPerWorld}{73.1}
\newcommand{\MeanCriteriaPerTask}{13.7}

\newcommand{\CandidatePoolSize}{599}
\newcommand{\QualifyingWorlds}{11}
\newcommand{\IncompleteTasksRemoved}{24}      
\newcommand{\TaskReauditRate}{5\%}           

\newcommand{\JudgeModel}{DeepSeek-v4-Flash}   
\newcommand{\JudgeSampleTasksTotal}{120} 
\newcommand{\JudgeSampleTasks}{40}

\newcommand{\JudgeCandidateCount}{10}         

\newcommand{\JudgeAnnotators}{3}
\newcommand{\InterAnnotatorAgreement}{92.8\%}  
\newcommand{\JudgeLabelN}{1{,}687}            
\newcommand{\JudgeCriteriaN}{1{,}687}
\newcommand{\JudgeConfusionTP}{799}
\newcommand{\JudgeConfusionFN}{21}
\newcommand{\JudgeConfusionFP}{28}
\newcommand{\JudgeConfusionTN}{839}
\newcommand{\JudgeAccuracy}{97.1\%}

\newcommand{\JudgePrecision}{96.6\%}
\newcommand{\JudgeRecall}{97.4\%}
\newcommand{\MetLabel}{Met}                  
\newcommand{\NotMetLabel}{Not Met}

\newcommand{\NPairwiseTests}{36}
\newcommand{\FDRLevel}{5\%}

\newcommand{\LoopHarness}{Loop Harness}
\newcommand{\RampHarness}{Ramp Harness}

\newcommand{\DollarBudgetOne}{\$1}
\newcommand{\DollarBudgetTwo}{\$5}
\newcommand{\DollarBudgetThree}{\$10}
\newcommand{\DollarBudgetFour}{\$50}

\begin{document}
\raggedbottom

\newgeometry{
  top=0.9in,
  bottom=0.8in,
  left=0.5in,
  right=0.5in,
  includehead,
  includefoot
}

\setlength{\headheight}{40pt}
\setlength{\headsep}{18pt}
\setlength{\parindent}{0pt}
\newcommand{\BannerHeader}{%
  \includegraphics[width=\headwidth,height=\headheight,keepaspectratio]{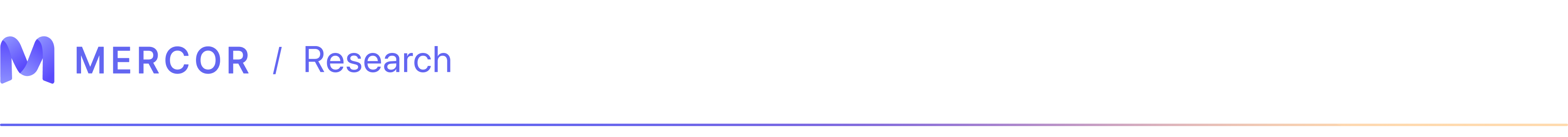}%
}

\pagestyle{fancy}
\fancyhf{}
\fancyhead[L]{\BannerHeader}
\fancyfoot[C]{\thepage}
\renewcommand{\headrulewidth}{0pt}

\fancypagestyle{plain}{
  \fancyhf{}
  \fancyhead[L]{\BannerHeader}
  \fancyfoot[C]{\thepage}
  \renewcommand{\headrulewidth}{0pt}
}

\thispagestyle{fancy}

\makeatletter
\twocolumn[
\begin{@twocolumnfalse}
\noindent
\vspace*{1.0cm}

{\huge APEX-Accounting\par}
\vspace{1.2cm}

{\large

Julien Benchek$^{\textcolor{mypurple}{1} *}$ \quad
Austin Bennett$^{\textcolor{mypurple}{1} *}$ \quad
Jasmin Kern$^{\textcolor{mypurple}{1} *}$ \quad
Ryan Stevens$^{\textcolor{darkgreen}{2}}$ \quad
Ren\'{e} Sultan$^{\textcolor{darkgreen}{2}}$ \quad
Charis Ching$^{\textcolor{mypurple}{1}}$ \quad
Hayley Popiel$^{\textcolor{mypurple}{1}}$ \quad
Vaibhav Mittal$^{\textcolor{mypurple}{1}}$ \quad
Felix Mercier$^{\textcolor{mypurple}{1}}$ \quad
Brendan Foody$^{\textcolor{mypurple}{1}}$ \quad
Bertie Vidgen$^{\textcolor{mypurple}{1}}$ \quad

}

\vspace{0.4cm}

{\large $^{\textcolor{mypurple}{1}}$Mercor \qquad $^{\textcolor{darkgreen}{2}}$Ramp\par}

\vspace{0.2cm}

{\small $^{*}$Equal contribution.\par}

\vspace{0.2cm}

{\small Contact \texttt{apex@mercor.com} for more information about \apex. \par}

\vspace{0.6cm}

\subsection*{Abstract}
We introduce \apex, a benchmark built by Mercor in partnership with Ramp, to assess whether frontier models can do the real work of accountants. Tasks include reconciling accounts, accruing expenses, posting transactions, and producing reports. The private eval set comprises $\NTasks{}$ tasks, split across $\NWorlds{}$ worlds. Each world contains an accounting system, as well as spreadsheets, PDFs, and other files. Every task was authored and solved by experts in accounting and bookkeeping, who also wrote grading rubrics. Across nine frontier models, \TopModel{} (Max) leads with \TopScore{} Mean Criteria@\KSubsample{}, ahead of Muse-Spark-1.1 (xHigh) at $52.6\%$. No model scores more than \MaxPassK{} Pass\textasciicircum{\NRuns{}} (GPT-5.6-Sol (Max+Pro)) and the highest Pass@\NRuns{} is $21.5\%$ (Muse-Spark-1.1 (xHigh)). We experiment with increasing the token budget from \DollarBudgetOne{} to \DollarBudgetFour{} and observe an instance of Simpson's paradox: scores increase as the token budget increases but within a given budget-constrained harness, scores are lower on tasks where the model spends more tokens. As \apex is a closed benchmark, leaderboard evals can be run for any frontier model on request.
\vspace{1.0cm}
\end{@twocolumnfalse}
]
\makeatother

\begingroup
\renewcommand{\thefootnote}{}
\renewcommand{\footnoterule}{}
\footnotetext{\footnotesize\textbf{Acknowledgments.} We thank all of the experts who contributed to the creation of \apex. We also thank Yash Bharti for his support in creating the eval set, Arnav Garg for his research assistance, and members of the Mercor research team for reviewing the paper.}
\endgroup

\section{Introduction}
Globally, accounting generates approximately $\$700$ billion~\citep{hughes2024takeaway} in revenue and the US alone employed close to 1.6 million accountants in 2024~\citep{onet2025accountants}. Producing correct end-of-year accounts and reporting tax correctly is a requirement for every company, regardless of their size or industry. This sort of knowledge work is skilled and detail-oriented, and requires deep understanding of business context while also being repetitive, procedural, and document-heavy. 
AI models have long passed the profession's certification exams: in 2023, GPT-4 with 10-shot prompting averaged $85.1\%$ across the CPA, CMA, CIA, and Enrolled Agent exams~\citep{eulerich2024hype}. A Stanford survey of $277$ accountants found that AI improved the quality of their work, despite them expressing concerns about data security and job stability~\citep{choi2025humanai}, and a 2026 Harvard Business School analysis scored accounting at 0.51 on a 0--1 scale of how readily generative AI can substitute human labor for a given occupation~\citep{chen2024displacement} and~\citep{azpurua2026enhance}.\\

We introduce \apex, a benchmark built by Mercor in partnership with Ramp. It comprises $\NTasks{}$ tasks across $\NWorlds{}$ synthetically generated worlds and spans four task types: \mbox{Reconciliation}, \mbox{Data Entry}, \mbox{Variance Analysis}, and \mbox{Schedules \& Accruals}. Each world is a self-contained company frozen at month-end close, and every task is authored and solved by accounting experts. Model outputs are graded by an LM judge against task-specific rubrics consisting of binary, outcome-based criteria. We also open-source a public dev set world ($n=\DevSetTasks{}$ tasks), including prompts, rubrics, and task
files.\footnote{\url{https://huggingface.co/datasets/mercor/apex-accounting}} \\

\begin{figure}[t]
\centering
\caption{Performance of models on the \apex{} leaderboard using Mean Criteria@\KSubsample{}, with $95\%$ task-bootstrap confidence intervals.}
\includegraphics[width=1\linewidth]{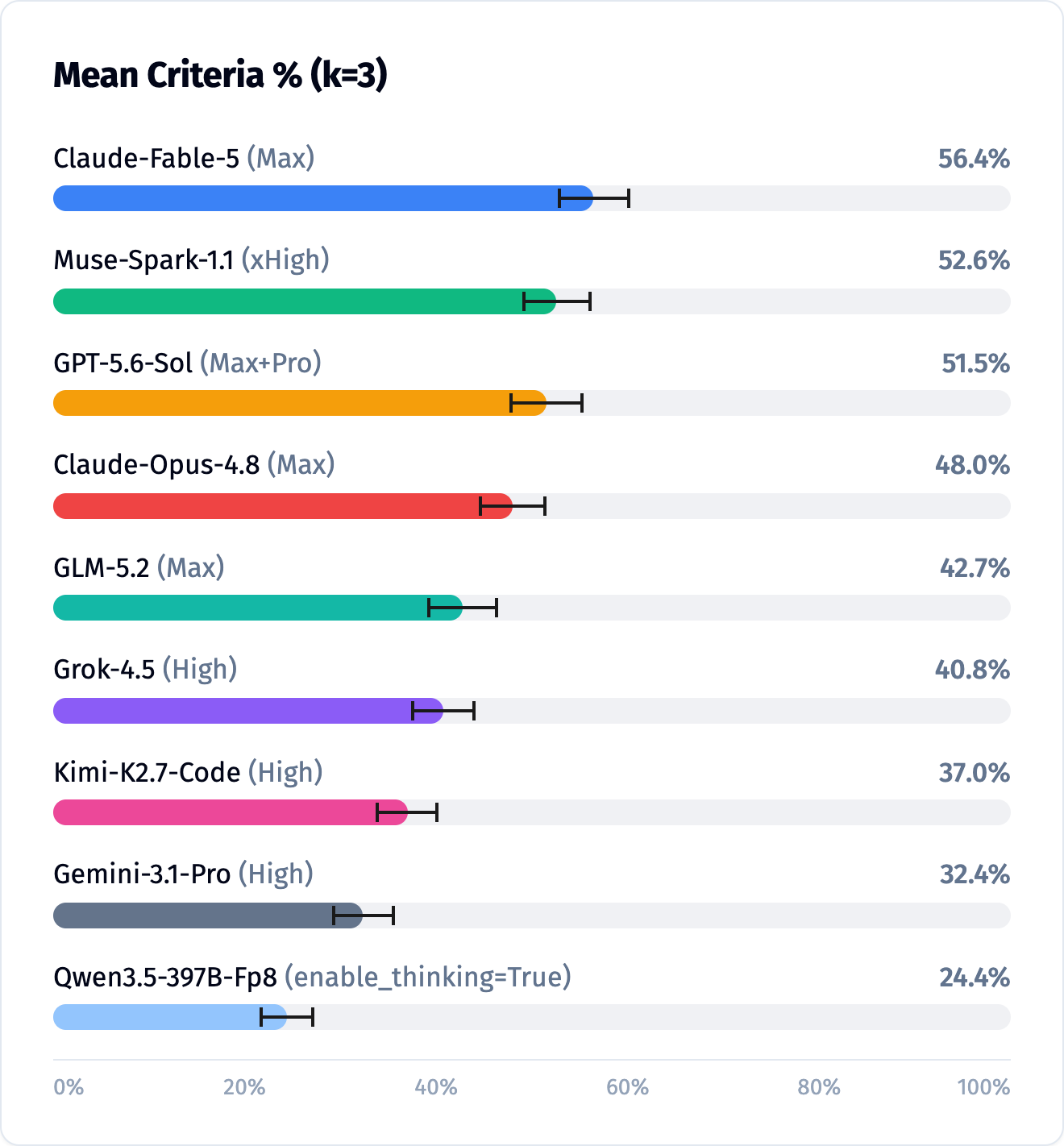}
\label{fig:leaderboard}
\end{figure}

Few benchmarks assess how AI models perform at real accounting work. AuditBench~\citep{wang2025auditbench} pairs real S\&P 500 financial statements with synthetic transactions to test inconsistency detection. FinMaster~\citep{jiang2025finmaster} spans the financial pipeline with text-only, simulator-generated tasks, and AccountingBench~\citep{penrose2025accountingbench} evaluates agents on closing the books for one real SaaS company against a human CPA. OpenAI's GDPval~\citep{patwardhan2025gdpval} has five tasks (of $n=220$) for accountants' and auditors' work in its open-sourced gold subset. None of these test the day-to-day bookkeeping and month-end close work that fill most of a bookkeeper or staff accountant's job at scale.\\

We evaluated $\NModels{}$ frontier models for the \apex{} leaderboard, using the Loop harness, which implements the canonical while-loop-with-tools agent architecture~\citep{goyal2025whileloop}. On Mean Criteria@\KSubsample{}, \TopModel{} is the best-performing model with \TopScore{}, followed by Muse-Spark-1.1 at $52.6\%$. The highest score for Pass@\NRuns{} is $21.5\%$ (Muse-Spark-1.1), and Pass\textasciicircum{\NRuns{}} at \MaxPassK{} (\ThirdModel{}).\\

We introduce a fine-grained taxonomy of failure categories, adapted to accounting work, and use it to annotate low-scoring trajectories from the three best-performing models. Their profiles are strikingly similar: reasoning failures dominate ($59$--$79\%$ of annotated failures per model). Information gathering and instruction following, which are often improved through better harnesses, account for only about a quarter combined. Two ablations point to the same: raising the per-task budget from \DollarBudgetOne{} to \DollarBudgetFour{} helps only models whose token appetite makes tight caps binding, and a purpose-built harness shifts Mean Criteria@\KSubsample{} by just $+1.2$\,pp on average. \\

\Cref{sec:benchmark_construction} describes how worlds, files, and tasks were designed by accounting experts. \Cref{sec:methodology} outlines our evaluation methodology, judge model, and metrics. \Cref{sec:results} reports the leaderboard, the budget ablations, and the harness experiments. \Cref{sec:failure_modes} reports top-performing models' failure modes to assist future model and agent development.

\begin{table*}[t]
\centering
\small
\setlength{\tabcolsep}{4pt}
\renewcommand{\arraystretch}{1.15}
\caption{\apex{} tasks by category ($n=\NTasks{}$), with category definitions. File counts are the files required to solve the prompt, whether drawn from the shared world corpus or provided with the task.}
\label{tab:dataset_stats}
\begin{tabularx}{\textwidth}{@{}l>{\RaggedRight}Xccccc@{}}
\toprule
\textbf{Category} & \textbf{Definition} & \textbf{Tasks} & \makecell{\textbf{Mean}\\\textbf{Criteria}} & \makecell{\textbf{Median}\\\textbf{Criteria}} & \makecell{\textbf{Mean}\\\textbf{Files}} & \makecell{\textbf{Median}\\\textbf{Files}} \\
\midrule
Reconciliation        & Tie two sources together, identify differences, and explain or correct the breaks.                    & $61$ & $14.61$ & $11$    & $7.49$ & $7$ \\
Data Entry            & Post or update transactions, journal entries, vendor bills, and invoices.                             & $26$ & $16.35$ & $13$    & $8.04$ & $6$ \\
Variance Analysis     & Compare actuals against budget, prior periods, or expectations and explain the drivers.               & $28$ & $14.54$ & $11$ & $7.86$ & $7$ \\
Schedules \& Accruals & Build or update supporting schedules, calculate accruals, and carry the right balances into the close. & $45$ & $10.29$ & $8$     & $7.00$ & $6$ \\
\midrule
Overall               & ---                                                                                                    & $\NTasks{}$ & $13.66$ & $10$ & $7.51$ & $7$ \\
\bottomrule
\end{tabularx}
\end{table*}

\section{Benchmark Design}
\label{sec:benchmark_construction}
\apex{} comprises $\NTasks{}$ tasks across $\NWorlds{}$ held-out synthetic company worlds, with $\TasksPerWorld{}$ tasks per world. The public dev set contains one world and $\DevSetTasks{}$ tasks. 

\subsection{Expert Selection}
\label{sec:expert_selection}

\apex{} was created by $\NExperts{}$ accounting experts with a median of $\ExpertMedianYears{}$ years of career experience (mean $\ExpertMeanYears{}$). Backgrounds skew toward experience in corporate finance and accounting ($35$ of $\NExperts{}$ experts). $\BigFourPct{}$ of experts have worked at a Big Four accounting firm (Deloitte, KPMG, EY, or PwC).

\subsection{World Construction}
Each world is a self-contained company, frozen at a specific month-end close. Each is based on a realistic business scenario and has its own entity type, chart of accounts, revenue model, and prior-period balances. Across their tasks, worlds draw on $\MeanFilesPerWorld$ unique required input files on average (see Table~\ref{tab:world_task_criteria_stats}). These were scoped in partnership with Ramp to ensure a diverse distribution of company profiles across revenue, headcount, industry, and geography. Every company follows U.S. GAAP accrual-basis accounting.\\

The worlds were built in four stages. First, we ran a cross-world scoping pass to assign each world a high-level profile, balanced across the full set. Second, we gave each world a detailed specification: a description of every file, every task to be built (including the deliberate traps each is meant to catch), and a trap register cataloging every seeded contradiction. Third, we iterated on a single reference document until its formatting and style were realistic, then derived a per-world style guide. We validated all files against the spec and style guide so that they read as coming from one company. Most world files were synthetically populated, always under expert specification and review. Every document is novel and screened against public sources, so no world can be found online or memorized in advance. On average, $\MeanExpertsPerWorld$ experts worked on each world.

\subsection{Task Creation}
Each task is associated with a single world. A task consists of a prompt, a set of required input files, a golden response, and a grading rubric.\footnote{\Cref{app:worked_example} shows one dev set task end-to-end --- the prompt, two representative rubric criteria, and a condensed evaluated trajectory --- to make the task and rubric structure concrete.} The required output for every task is a message sent in the console. On average, there are $\MeanFilesPerTask{}$ required input files per task that the model needs to find and read. Most of these belong to the shared world file corpus, but about half of the tasks provide additional task-specific files, which are placed in the filesystem alongside the world files when the task begins. Across the held-out set, $83$ of $\NTasks{}$ tasks ($52\%$) include at least one task-specific file (median $2$, mean $2.7$ per such task). Task-specific files are most common in Reconciliation ($33/61$ tasks), Schedules \& Accruals ($29/45$), and Variance Analysis ($17/28$), and rare in Data Entry ($4/26$). \\

Prompts mimic realistic requests that an accountant would receive on the job. They are concise and include a clear expectation of the final output but do not explain the methods where a competent accountant would already know them, nor give the file names unless they are unintuitive or unexpected.\footnote{Quality control processes applied during dataset construction, including expert review, automated QC, and independent re-solve baselining, are given in \Cref{app:quality_control}.} Each task is in one of four categories: Reconciliation, Data Entry, Variance Analysis, and Schedules \& Accruals. \Cref{tab:dataset_stats} defines each category and reports its task count and per-task criteria and file statistics. \\

Experts created a grading rubric for each task, comprising binary (Pass/Fail) unweighted criteria. On average, there are $\MeanCriteriaPerTask{}$ criteria per task. To ensure grading fairness, each criterion is self-contained, easy to interpret, aligned with the prompt, and outcome-based (\Cref{tab:criterion_requirements}). Experts also created a golden response for each task, which represents a top-tier industry-quality output, scoring $100\%$ against the rubric. \\

\begin{table}[t]
\centering
\small
\renewcommand{\arraystretch}{1.15}
\caption{Requirements every rubric criterion must satisfy.}
\label{tab:criterion_requirements}
\begin{tabularx}{\columnwidth}{@{}l>{\RaggedRight}X@{}}
\toprule
\textbf{Requirement} & \textbf{Definition} \\
\midrule
Self-contained      & Gradable without reference to other criteria. \\
\addlinespace[3pt]
Easy to interpret   & One fact or judgment per criterion, so partial credit is never lost to compound requirements. \\
\addlinespace[3pt]
Aligned with prompt & Grading only what the prompt asks. \\
\addlinespace[3pt]
Outcome-based       & Grading only the final answer, not intermediate steps. We apply acceptable ranges to handle legitimate rounding differences. \\
\bottomrule
\end{tabularx}
\end{table}

\section{Experimental Setup}
\label{sec:methodology}

\subsection{Models}
We evaluate $\NModels{}$ frontier models for the \apex{} leaderboard: \TopModel{}, Muse-Spark-1.1, \ThirdModel{}, \SecondModel{}, GLM-5.2, Grok-4.5, Kimi-K2.7-Code, Gemini-3.1-Pro, and Qwen3.5-397B-Fp8. Each model independently executes each task $\NRuns{}$ times, resulting in a total of $\NTrajectories{}$ trajectories. A trajectory comprises the sequence of steps a model takes, where each step is one model turn. It can include output tokens, reasoning tokens, and one or more tool calls. We allow a maximum of $500$ steps and $5$ million tokens per task.\footnote{Provider, context window, maximum output, thinking-effort configuration, and wall-clock time per run for every model are reported in \Cref{app:model_details}.}

\subsection{Harnesses}
\label{sec:harness}
\begin{figure*}[t]
\centering
\caption{\RampHarness{} schematic.}
\includegraphics[width=\textwidth]{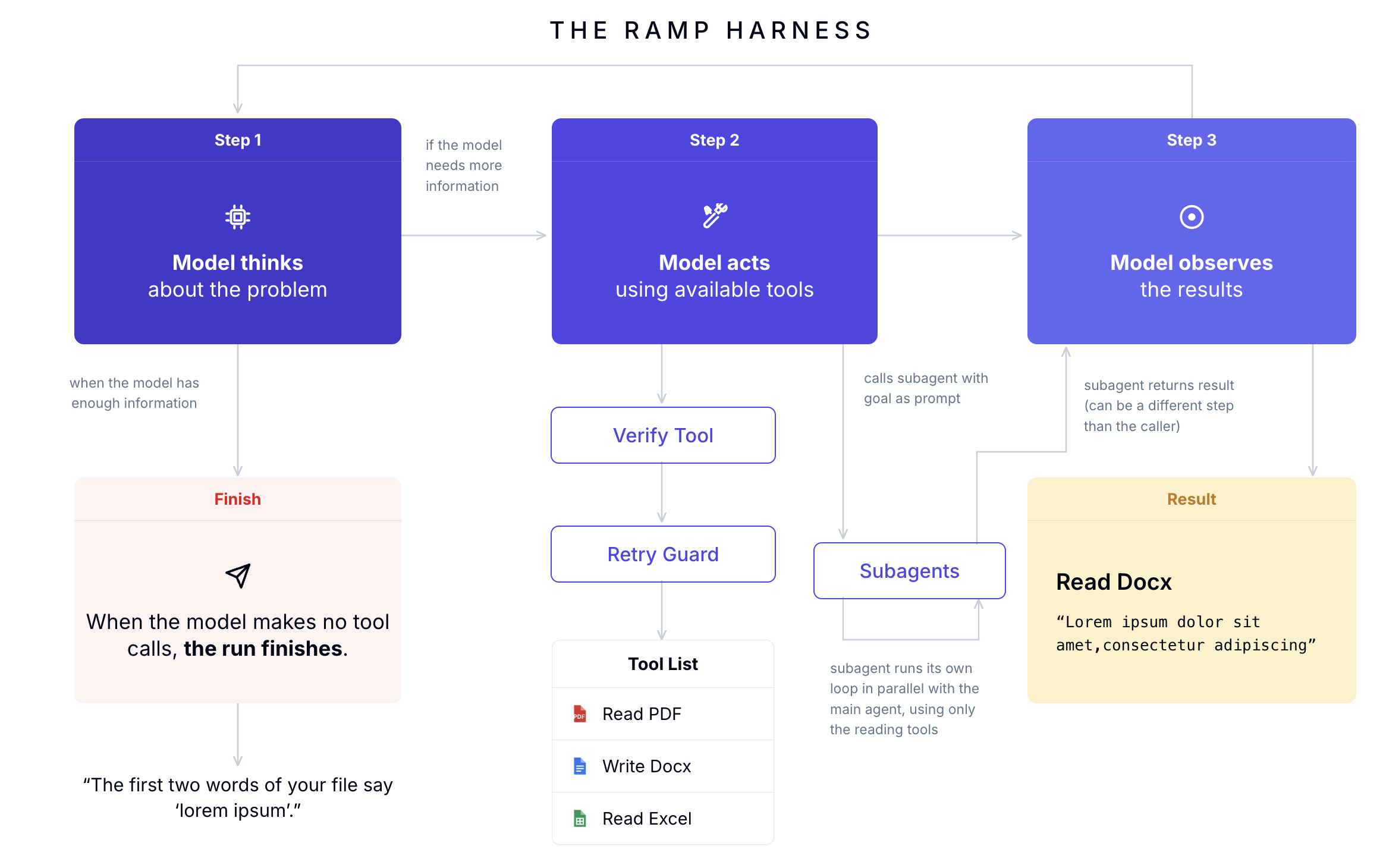}
\label{fig:harness_loops}
\end{figure*}

We evaluate models using two harnesses: a standard \LoopHarness{} and the \RampHarness{}. Both harnesses are implemented and executed within Archipelago, our internal framework for running agent evaluations at scale.\footnote{\url{https://github.com/Mercor-Intelligence/archipelago}}
Leaderboard results use the \LoopHarness{}. The \LoopHarness{} repeatedly passes context into a model to solve a task. The \RampHarness{}, however, expands on a ReAct-style harness ~\citep{yao2023react} with parallel subagents. This \RampHarness{} was built in partnership with Ramp to mirror their internal harness infrastructure. We do not include any specialized skills or self-learning loops discussed in a prior accounting agent analysis ~\citep{stevens2026stackbenchmarking}. Thus, we are testing performance under different agent runtime environments, not Ramp's accounting agentic product itself. \Cref{fig:harness_loops} shows a schematic of the \RampHarness{}. Specifically, there are six properties designed to mirror the constraints of a real accounting deployment: \\

\begin{enumerate}[nosep, leftmargin=1.6em, itemsep=2pt, topsep=2pt]
  \item \textit{Retry-awareness:} Before executing a tool call, the harness checks whether an identical call has already been made; if so, it blocks the call, avoiding infinite loops and inefficient repeated tool use.
  \item \textit{Tool allowlist:} Each agent is exposed to only the set of tools needed to solve the problem.
  \item \textit{Tool validation:} Tool arguments are parsed and validated before execution.
  \item \textit{Unavailable-tool recovery:} If the model requests a tool it cannot use, the harness returns a model-visible tool error rather than executing anything.
  \item \textit{Read/write execution policy:} Read-only tools can run in parallel through subagents, while write tools run serially to avoid conflicting side effects.
  \item \textit{Subagent delegation:} Some tool calls can launch a separate agent loop to complete a narrower sub-task, then return its result to the parent loop.
\end{enumerate}


\vspace{0.5em}
For the canonical leaderboard evals we restrict agents to $5$ million tokens and $500$ steps. We created versions of the harnesses that tell the model at each step how many steps and tokens remain. If the step or token limit is reached, the model is told to submit a final answer; failure to do so is scored zero. If the token limit is reached we alert the model that the budget is exhausted and give it one additional turn to submit the answer. Any subagents that are currently active are canceled without being allowed to return. This ensures graceful stopping without a significant advantage beyond the original token budget. \\

Additionally, we conducted an experiment for \TopModel{}, \ThirdModel{}, and Gemini-3.1-Pro, where we compared unlimited token spend against the $5$ million token cap. For all three models, the $5$ million token limit scores were slightly higher, supporting this design choice. The harness reports remaining tokens at every step, and a finite, visible budget pushes the model to consolidate intermediate results and commit to an answer, whereas unlimited runs keep exploring past their best candidate and may dilute the final submission with accumulated context.


\subsection{Tool Use}
Both harnesses were given $91$ operational tools to choose from: filesystem, docx, pdf, excel, mail, code execution, and accounting software. Additionally, the \RampHarness{} was given $11$ meta-tools to facilitate the run (e.g. subagent operations). Operational tools aid the model's analysis through read and write functions. The first six categories are standard tool families for agent harnesses, while the accounting-software tools supplement them, offering a lightweight version of common accounting operations.

\subsection{Grading Methodology}
\JudgeModel{}, configured with temperature at $0.1$, high reasoning, and a GEPA-optimized~\citep{gepa} grading prompt, is the grading judge. To grade each criterion the judge receives the task prompt, the criterion text, and the model's final output, but not the trajectory log. It returns a binary score (\MetLabel{}, \NotMetLabel{}) along with a concise free-text explanation. \\

To select \JudgeModel{} as the grading judge, we sampled $\JudgeSampleTasks{}$ trajectories from \TopModel{}, \ThirdModel{}, and Gemini-3.1-Pro. This resulted in a total of $\JudgeSampleTasksTotal{}$ trajectories, with $\JudgeCriteriaN$ criteria. Three experts independently annotated every criterion for each trajectory, and we took the majority vote across the $\JudgeAnnotators{}$ annotators as ground truth for that criterion; raw inter-annotator agreement is $\InterAnnotatorAgreement{}$, with $89.2\%$ of criteria labeled unanimously. Fleiss' $\kappa = 0.857$, indicating strong agreement beyond chance.\footnote{\Cref{tab:interrater} (\Cref{app:judge_eval}) reports the full inter-rater reliability values.} \\

We used the ground truth dataset to assess $\JudgeCandidateCount$ LM judges, including $7$ frontier models (Claude-Opus-4.8 $[$reasoning = high/low$]$, Gemini-3.1-Pro $[$reasoning = high/low$]$, Gemini-3-Flash, GPT-5.5 $[$reasoning = high/low$]$) and $3$ open-weight models (Kimi-K2.6, DeepSeek-v4-Flash [reasoning = high], DeepSeek-v4-Pro [reasoning = high]). \JudgeModel{} was selected because its performance is within noise of the best frontier judge, Claude-Opus-4.8 (High) ($F_1 = 0.970$ for both), while being open-weight. \Cref{tab:judge_confusion_matrix} shows a confusion matrix for the $\JudgeLabelN{}$ majority-vote ground truth labels. The judge achieves $\JudgeAccuracy{}$ overall accuracy, with $\JudgePrecision{}$ precision and $\JudgeRecall{}$ recall on the \MetLabel{} class.\footnote{\Cref{app:judge_eval} reports judge agreement broken out by solver model and shows no per-model inflation; the full comparison across all $\JudgeCandidateCount{}$ candidate judges is reported in \Cref{tab:llmaaj_scores} (\Cref{app:judge_eval}).}

\begin{table}[t]
\centering
\small
\caption{Confusion matrix for the judge model (\JudgeModel{}), evaluated against majority-vote human ground truth ($n=\JudgeCriteriaN{}$ criteria, one majority-vote label per criterion).}
\label{tab:judge_confusion_matrix}
\begin{tabular}{lcc}
\toprule
 & Predicted: \MetLabel{} & Predicted: \NotMetLabel{} \\
\midrule
Actual: \MetLabel{}    & $\JudgeConfusionTP{}$ & $\JudgeConfusionFN{}$   \\
Actual: \NotMetLabel{} & $\JudgeConfusionFP{}$   & $\JudgeConfusionTN{}$ \\
\bottomrule
\end{tabular}
\end{table}

\subsection{Metrics}
We report four metrics: Mean Criteria@\KSubsample{}, Pass@1, Pass@k, and Pass\textasciicircum{k} (for $k = 1$ through $\NRuns{}$). Mean Criteria@\KSubsample{} is our primary leaderboard metric: the percentage of rubric criteria met, averaged across $\KSubsample{}$ of a model's $\NRuns{}$ runs per task, chosen lexicographically based on task ID strings, which essentially make up a random sample. We use Mean Criteria@\KSubsample{} because per-criterion partial credit separates models that complete most of a task from models that fail outright. Averaging over $\KSubsample{}$ runs rather than all $\NRuns{}$ reduces run-to-run variance at a lower computational cost.\footnote{Runs that fail because of the model are scored as zero, while runs that fail for reasons outside the model's control are excluded from every metric. \Cref{app:run_completion} gives the full convention and its effect on task coverage.}\\

Beyond the leaderboard, we report Pass@\NRuns{} as an indicative ceiling on current frontier agent capabilities. It measures whether a model passes a task at least once across $\NRuns{}$ attempts. We also report Pass\textasciicircum{\NRuns{}} \citep{yao2024tau}, which measures whether a model produces a fully correct output in each of its $\NRuns{}$ attempts. We compute $95\%$ confidence intervals via task-level bootstrapping, resampling individual tasks with replacement across $10{,}000$ resamples.

\begin{figure}[t]
\centering
\caption{Pass@k versus Pass\textasciicircum{k} by model, $k=1$ through $\NRuns{}$, capability ceiling versus consistency floor. Claude-Fable-5 (Max) and Muse-Spark-1.1 (xHigh) scores are bolded for visibility.}
\includegraphics[width=\columnwidth]{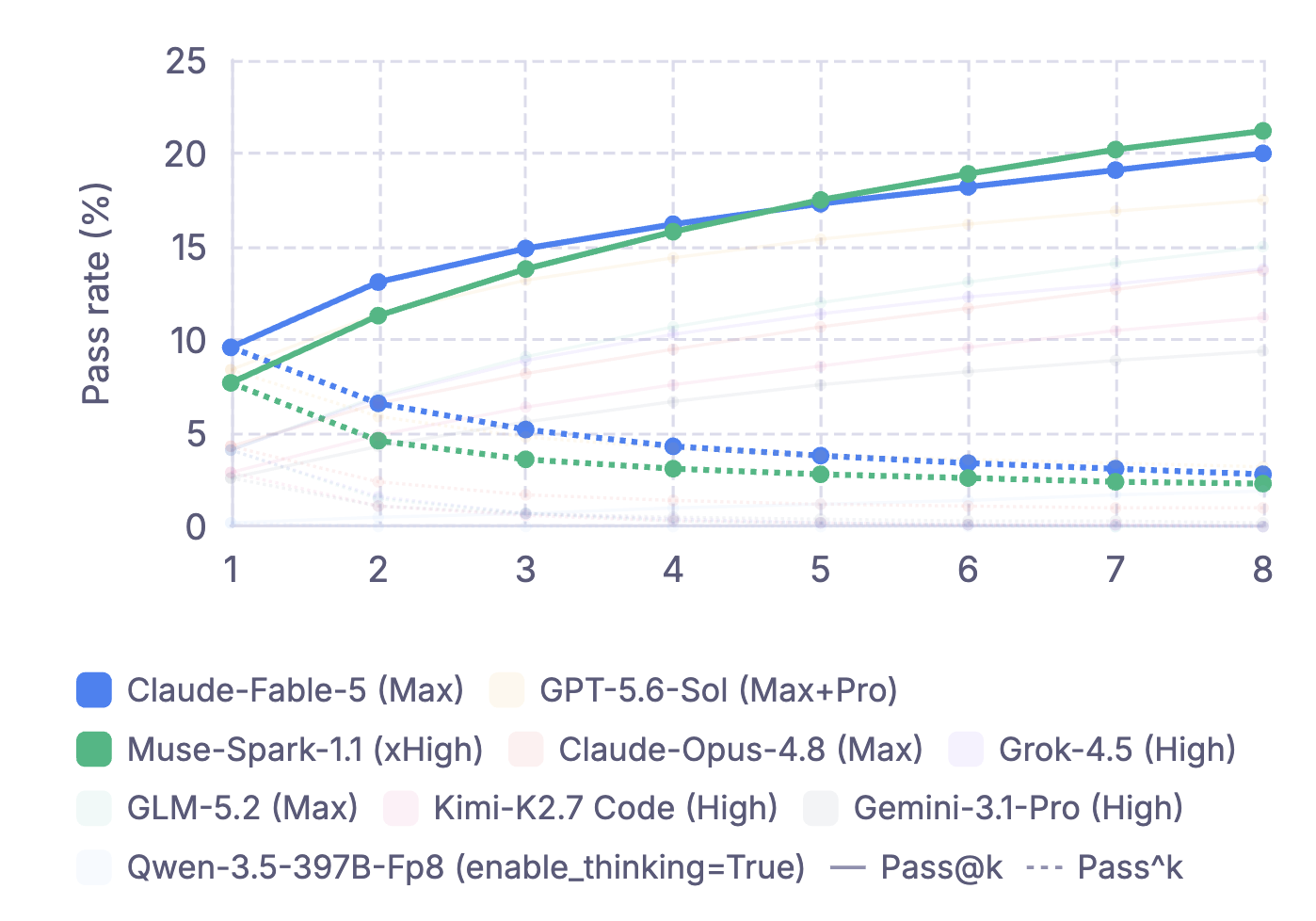}
\label{fig:pass_k_vs_passhat_k}
\end{figure}

\begin{table*}[t]
\centering
\caption{Mean Criteria@\KSubsample{} by task category, for all $\NModels{}$ models.}
\label{tab:task_category}
\begin{tabular}{lcccc}
\toprule
\textbf{Model} & \textbf{Reconciliation} & \textbf{Data Entry} & \textbf{Variance Analysis} & \textbf{Schedules \& Accruals} \\
\midrule
\textit{Tasks} ($N$) & $61$ & $26$ & $28$ & $45$ \\
\midrule
Claude-Fable-5 (Max)          & $59.4\%$ & $58.6\%$ & $56.8\%$ & $51.0\%$ \\
Muse-Spark-1.1 (xHigh)        & $52.8\%$ & $57.9\%$ & $57.2\%$ & $46.4\%$ \\
GPT-5.6-Sol (Max+Pro)         & $55.3\%$ & $52.4\%$ & $50.2\%$ & $46.5\%$ \\
Claude-Opus-4.8 (Max)         & $53.2\%$ & $48.7\%$ & $51.9\%$ & $38.1\%$ \\
GLM-5.2 (Max)                 & $44.1\%$ & $50.0\%$ & $51.7\%$ & $31.0\%$ \\
Grok-4.5 (High)               & $47.4\%$ & $38.4\%$ & $45.5\%$ & $30.2\%$ \\
Kimi-K2.7-Code (High)         & $40.5\%$ & $40.1\%$ & $44.3\%$ & $25.9\%$ \\
Gemini-3.1-Pro (High)         & $34.6\%$ & $28.9\%$ & $36.2\%$ & $28.9\%$ \\
Qwen3.5-397B-FP8 (True)       & $28.5\%$ & $20.9\%$ & $24.9\%$ & $20.6\%$ \\
\bottomrule
\end{tabular}%
\end{table*}

\section{Results}
\label{sec:results}
\subsection{Leaderboard}
\label{sec:leaderboard}
Figure~\ref{fig:leaderboard} shows the full APEX\textendash Accounting leaderboard.\footnote{\Cref{tab:leaderboard} (\Cref{app:devset_shift}) reports each model's Mean Criteria@\KSubsample{}, Pass@1, Pass@\NRuns{}, and Pass\textasciicircum{\NRuns{}} in full, with $95\%$ bootstrap confidence intervals.} \TopModel{} performs best at \TopScore{}, ahead of Muse-Spark-1.1 ($52.6\%$); \ThirdModel{} (\ThirdScore{}) and \SecondModel{} (\SecondScore{}) follow close behind. The remaining models---GLM-5.2, Grok-4.5, Kimi-K2.7-Code, Gemini-3.1-Pro, and Qwen3.5-397B---trail, with Qwen3.5-397B lowest at $24.4\%$. \\

Because the leaderboard contains $\NModels{}$ models, we conduct $\NPairwiseTests{}$ pairwise significance tests; we control the false discovery rate at $\FDRLevel{}$ with the Benjamini\textendash Hochberg procedure over per-task Mean Criteria@\KSubsample{} differences.\footnote{\Cref{app:pairwise} reports the full pairwise comparison matrix and which leaderboard gaps remain significant after correction using paired t-tests. \Cref{app:devset_shift} compares each model's leaderboard performance against the public dev set world.} All gaps spanning two or more leaderboard ranks remain significant after correction; only two adjacent-rank pairs are statistically indistinguishable. \\

\begin{figure*}[t]
\centering
\caption{Panel A. For five models, the Mean Criteria@\KSubsample{} versus the per-task dollar budget cap (set at \DollarBudgetOne{}, \DollarBudgetTwo{}, \DollarBudgetThree{} and \DollarBudgetFour{}). Panel B. The Mean Criteria@\KSubsample{} versus the mean total tokens used at each budget cap. At higher budget caps models do not, on average, fully utilize the tokens available to them.}
\includegraphics[width=\textwidth]{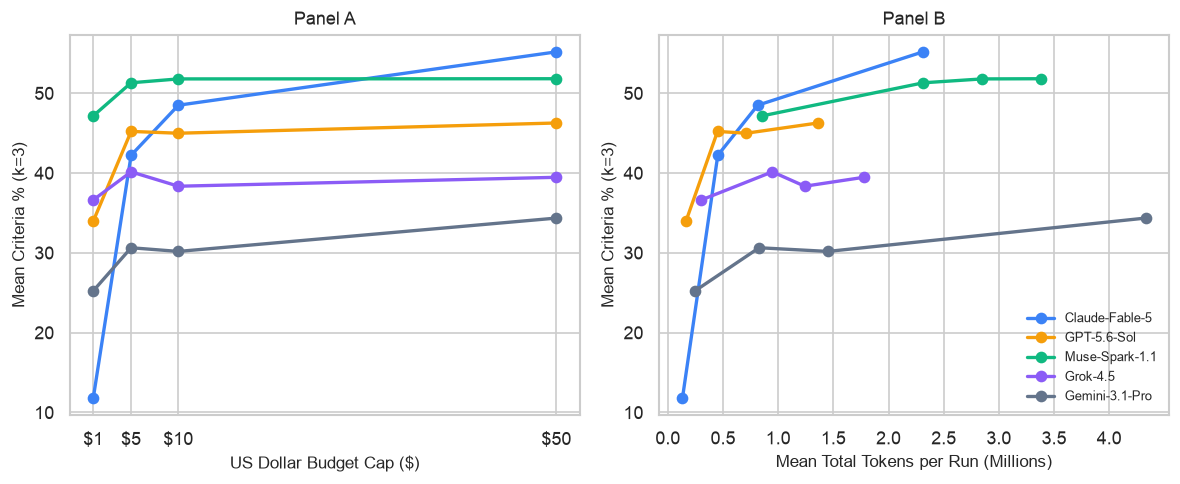}
\label{fig:budget_ablation}
\end{figure*}

\subsection{Performance by Task Category}
\Cref{tab:task_category} breaks down each model's Mean Criteria@\KSubsample{} by question category (given in \Cref{tab:dataset_stats}). Claude-Fable-5 leads three of the four categories (Reconciliation, Data Entry, and Schedules \& Accruals) while Muse-Spark-1.1 edges ahead on Variance Analysis ($57.2\%$ vs.\ $56.8\%$). It sits within a point of Claude-Fable-5 on Data Entry, consistent with the two models' close aggregate scores. Schedules \& Accruals is the hardest category, with every model scoring lowest (Gemini-3.1-Pro ties its Data Entry score) and trailing their highest-scoring categories by $7$--$21$ percentage points, reflecting the category's multi-step, judgment-heavy nature. \\

\subsection{Pass@\NRuns{} and Pass\textasciicircum{\NRuns{}}}
Pass@\NRuns{} and Pass\textasciicircum{\NRuns{}} measure models' practical capability ceiling and consistency, respectively. The capability ceiling shows the potential maximum performance of models, whereas consistency is often the most important consideration for real-world deployment of AI models. Figure~\ref{fig:pass_k_vs_passhat_k} plots Pass@k against Pass\textasciicircum{k} for $k=1$ through $\NRuns{}$. The two metrics rank models differently: Muse-Spark-1.1 achieves the highest Pass@\NRuns{} ($21.5\%$), narrowly ahead of \TopModel{} (\TopPassEight{}), whereas \ThirdModel{} scores the highest Pass\textasciicircum{\NRuns{}} at \MaxPassK{}. Consistency is low across the board, no model exceeds \MaxPassK{} on Pass\textasciicircum{\NRuns{}}, demonstrating how far agents are from reliably completing accounting work end to end. \\

\subsection{Cost Ablation Experiments}
\label{sec:token_ablation}
We tested five models---\TopModel{}, \ThirdModel{}, Gemini-3.1-Pro, Grok-4.5, and Muse-Spark-1.1---using the \LoopHarness{} and Mean Criteria@\KSubsample{} at dollar token budgets of \DollarBudgetOne{}, \DollarBudgetTwo{}, \DollarBudgetThree{}, and \DollarBudgetFour{} per task. Standardizing on cost accounts for the different ways that models price and tokenize. To calculate the budget per model, we assume a $90$:$10$ split of input to output tokens using the higher cost tier listed by model providers (prices can vary by token count).\footnote{\Cref{app:cost_conversion} details this per-model dollar-to-token conversion (\Cref{tab:cost_conversion}) and shows an example system-prompt excerpt disclosing the resulting token budget to the model (\Cref{fig:cost_prompt_example}).} The harness tells the model at each step an updated estimate of its remaining tokens. \\

Because every task is run at every budget, we measure each model's gain as a paired per-task difference, which removes task difficulty from confounding the comparison.\footnote{\Cref{tab:paired_deltas} reports all paired budget contrasts, with Benjamini--Hochberg-corrected significance.}
Raising the dollar budget leads to higher scores, but the impact differs sharply across models (\Cref{fig:budget_ablation}). 
The impact is greatest on models whose high token cost means that they have very few tokens at the lower budgets, such as Claude-Fable-5. \TopModel{} scores $11.8\%$ at \DollarBudgetOne{}---scoring exactly zero on $38.8\%$ of its tasks, more than triple any other model---and increases to $55.2\%$ at \DollarBudgetFour{}. This is a gain of $+43.4$~pp, with significant improvements at every budget step. Muse-Spark-1.1 is the opposite case: it leads at the tightest cap ($47.2\%$) yet gains only $+4.7$~pp across the whole range. The consequence is that at \DollarBudgetFour{}, \TopModel{} spends \$32.36 per run on average against Muse-Spark-1.1's \$5.25, yet their mean criteria scores are within 4 percentage points. \\

Outside of the \DollarBudgetOne{} cap, models typically do not fully utilize the token limits. There is only a moderate increase in token usage when going from \DollarBudgetThree{} to \DollarBudgetFour{} and, on average, models utilize just $64.7\%$ of the maximum budget available at \DollarBudgetFour{}. This cap constrains just $0\%$ to $16.9\%$ of runs across the five models.\footnote{\Cref{app:spend_score_slope} reports the per-cell within-cap slopes, \Cref{tab:panel_b_9010} reports the within-cap levels slope $b$ (change in Mean Criteria@\KSubsample{} per dollar of realized spend) for each model$\times$budget cell.} \\

\begin{figure*}[t]
\centering
\caption{Per-task Mean Criteria@\KSubsample{} versus realized spend per run (USD), by model, at each dollar budget ($90$:$10$ input:output split). Lines are per-model within-cap OLS fits, annotated with the slope $b$.}
\includegraphics[width=\textwidth]{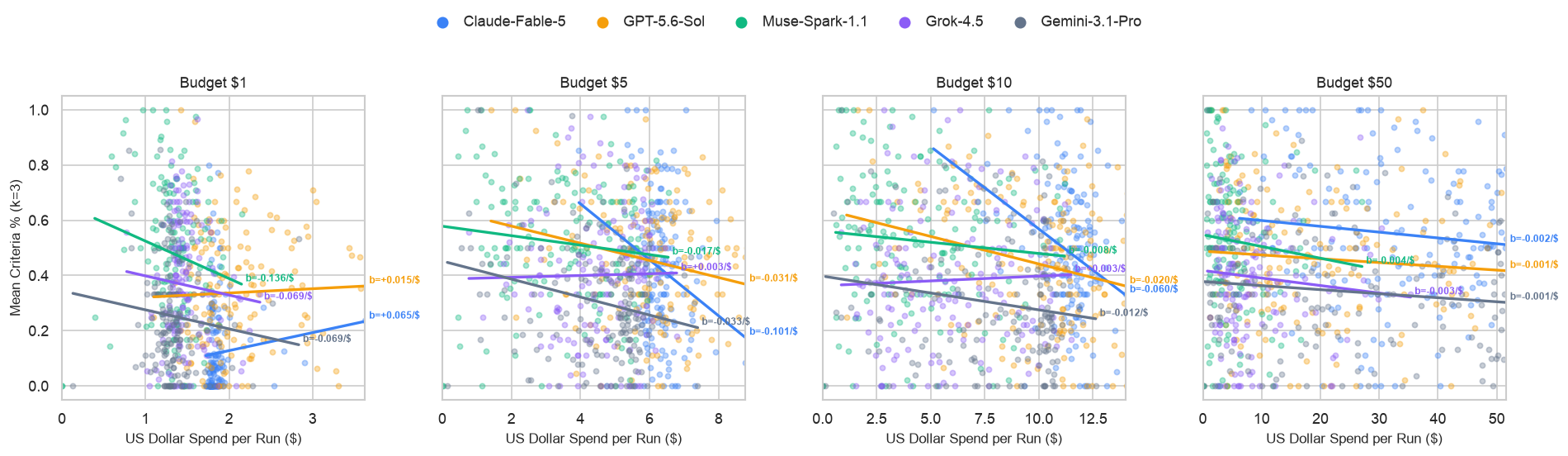}
\label{fig:spend_score_panels}
\end{figure*}

\begin{figure*}[t]
\centering
\caption{Task-level change in Mean Criteria@\KSubsample{} relative to the \DollarBudgetFour{} baseline, versus realized spend per run, by model. Points are colored by budget cap (\DollarBudgetOne{}, \DollarBudgetTwo{}, \DollarBudgetThree{}); black lines are pooled OLS fits, with per-cap slopes $b$ annotated above each panel. X-axis and Y-axis truncated to middle 98\% of values for visual clarity.}
\includegraphics[width=\textwidth]{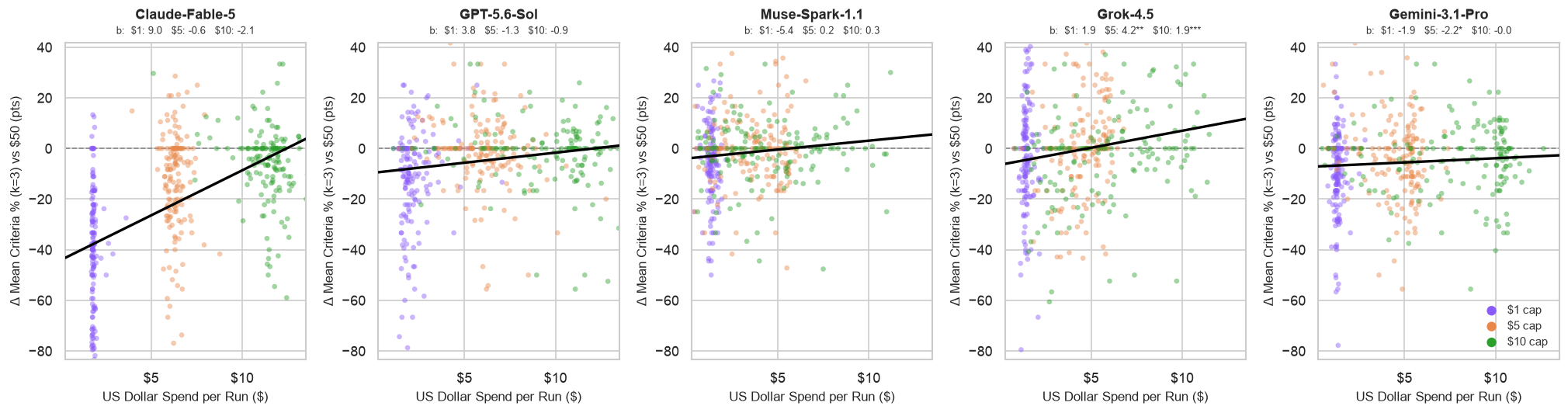}
\label{fig:delta_vs_50_baseline}
\end{figure*}

Within a fixed budget, tasks where models use more tokens are not associated with higher scores. After Benjamini--Hochberg correction, every statistically significant within-cap-spend score association is negative. This presents an example of Simpson's Paradox, wherein raising the budget raises a model's scores, yet within any fixed budget harness, the tasks on which a model spends more tokens score lower. This is because task difficulty confounds the result. Harder tasks use more tokens but still score lower. \Cref{fig:spend_score_panels,fig:delta_vs_50_baseline} show the per-task spend--score scatter with fitted within-cap slopes and the slope summary. \\

\subsection{Harness Experiments}
\label{sec:harness_experiments}

\begin{figure}[t]
\centering
\caption{Mean Criteria@\KSubsample{} by model under the \LoopHarness{} versus the \RampHarness{} ($k=\KSubsample{}$, $n=\NTasks{}$). Error bars are task-bootstrap $95\%$ CIs; \ThirdModel{} has no \RampHarness{} run.}
\includegraphics[width=\columnwidth]{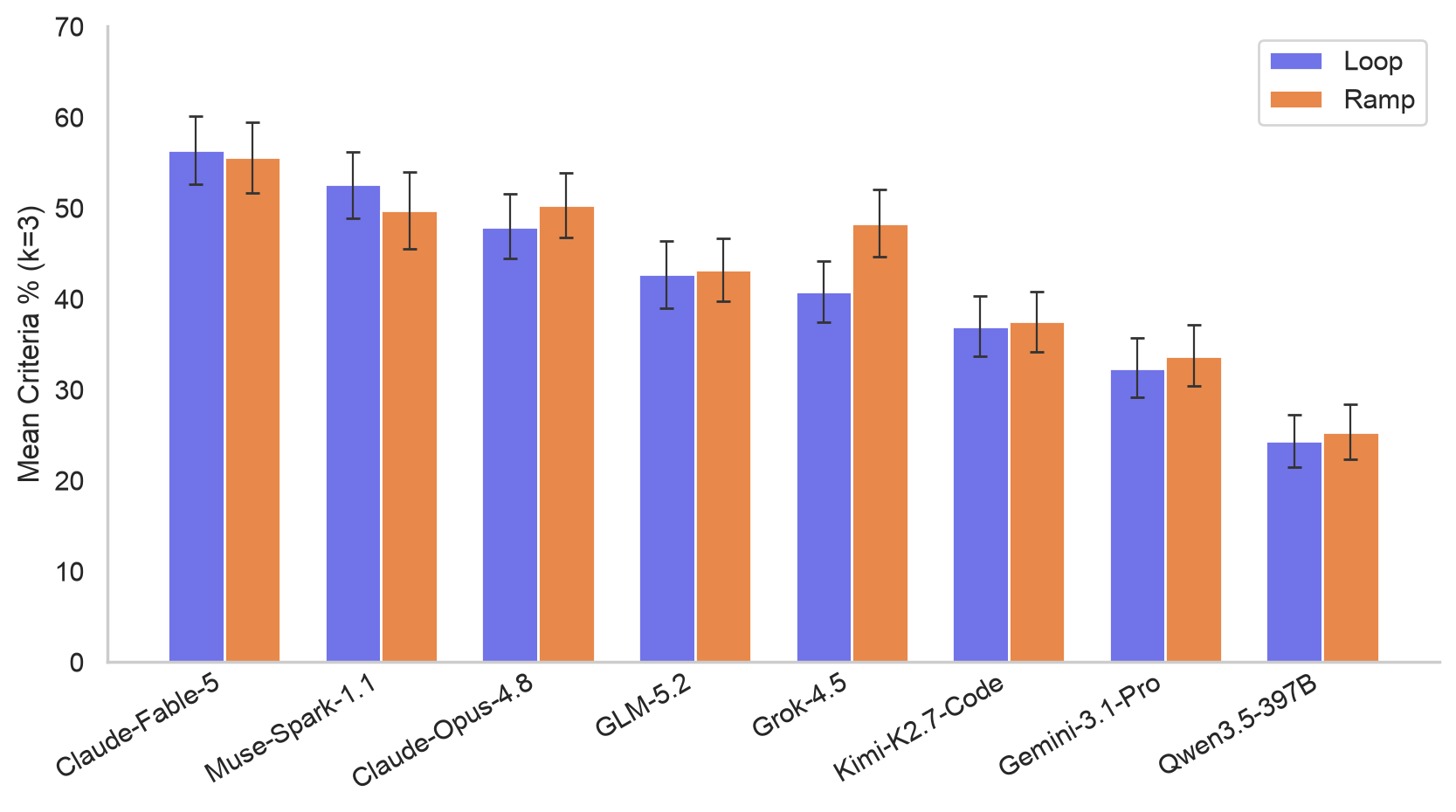}
\label{fig:harness_crit3}
\end{figure}

We compare performance of the \LoopHarness{} and the \RampHarness{} for each model. Harness choice does not have a substantial impact on Mean Criteria@\KSubsample{}. This implies that between the two harness architectures  we tested, the runtime environment has little performance impact. Averaged over the eight models run under both harnesses, the raw result shift (Ramp $-$ Loop) is $+1.2$\,pp, and just $+0.3$\,pp excluding one outlier. Grok-4.5's gain is the only shift that survives multiple-testing correction; for every other model, the per-harness CIs overlap. Grok-4.5 gains $+7.5$\,pp ($40.8$ to $48.3$), overtaking GLM-5.2 to place fourth among the paired models. Ramp's harness does not systematically favor the strongest models either: Claude-Fable-5 and Muse-Spark-1.1 both score \emph{lower} under the \RampHarness{} ($-0.8$ and $-2.8$\,pp), while Claude-Opus-4.8 gains $+2.4$\,pp and edges past Muse-Spark-1.1 (based on raw table difference). Switching harness is therefore not a free performance lift: on \apex{} the model had more of an impact than the underlying harness. Harness scores are shown in \Cref{fig:harness_crit3}.\footnote{\Cref{app:harness_details} reports the full comparison, including Pass@1 and corrected significance for every paired contrast.} \\

Only the \RampHarness{} allows subagent-delegation tools, and models differ sharply in how much they use them---from \TopModel{} ($4.9$ subagents per task, $89\%$ of trajectories) to Gemini-3.1-Pro and Qwen3.5-397B ($0.2$--$0.4$ per task, $12$--$13\%$). Delegation propensity does not line up with harness gains: GLM-5.2 delegates nearly as often as \TopModel{} yet both shift by less than a point, while the one significant \RampHarness{} improvement (Grok-4.5) comes from a model that delegates in $69\%$ of its trajectories. Delegation propensity reads more as a model trait than a mechanism that raises scores.\footnote{\Cref{app:subagent} reports per-model spawn rates.}

\begin{figure*}[t]
\centering
\caption{L1 failure-mode counts for the three best-performing models, over annotated failures ($n=24$/$28$/$22$ for Claude-Fable-5 (Max), GPT-5.6-Sol (Max+Pro), and Muse-Spark-1.1 (xHigh)). Reasoning failures dominate every profile; no annotated failure involves tool use.}
\includegraphics[width=0.9\textwidth]{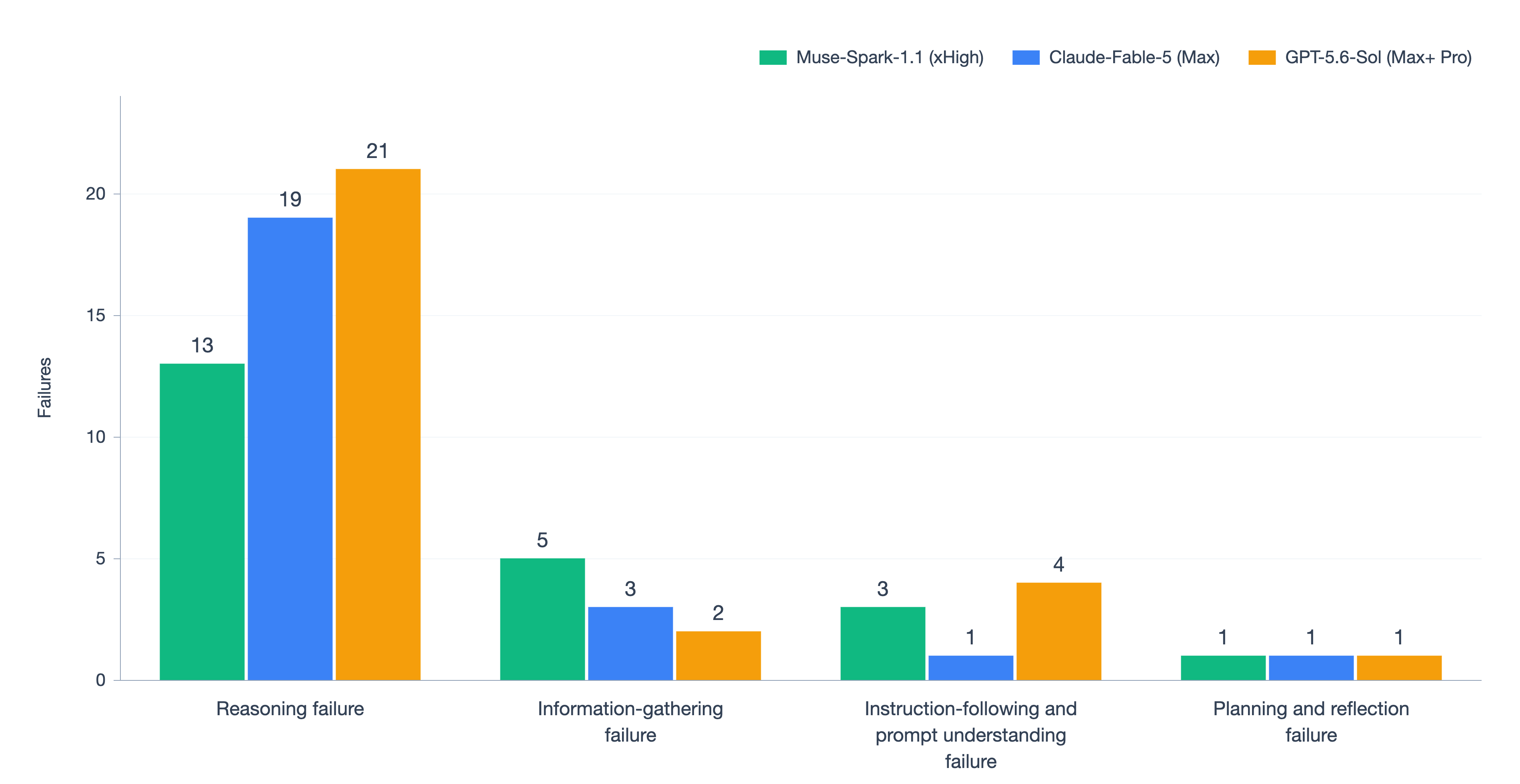}
\label{fig:failure_modes_top3}
\end{figure*}

\begin{figure*}[t]
\centering
\caption{L2 subclass counts within Reasoning failure for the three best-performing models. Non-numeric reasoning failures and data handling errors dominate; forgetting the information recurs across all three.}
\includegraphics[width=0.9\textwidth]{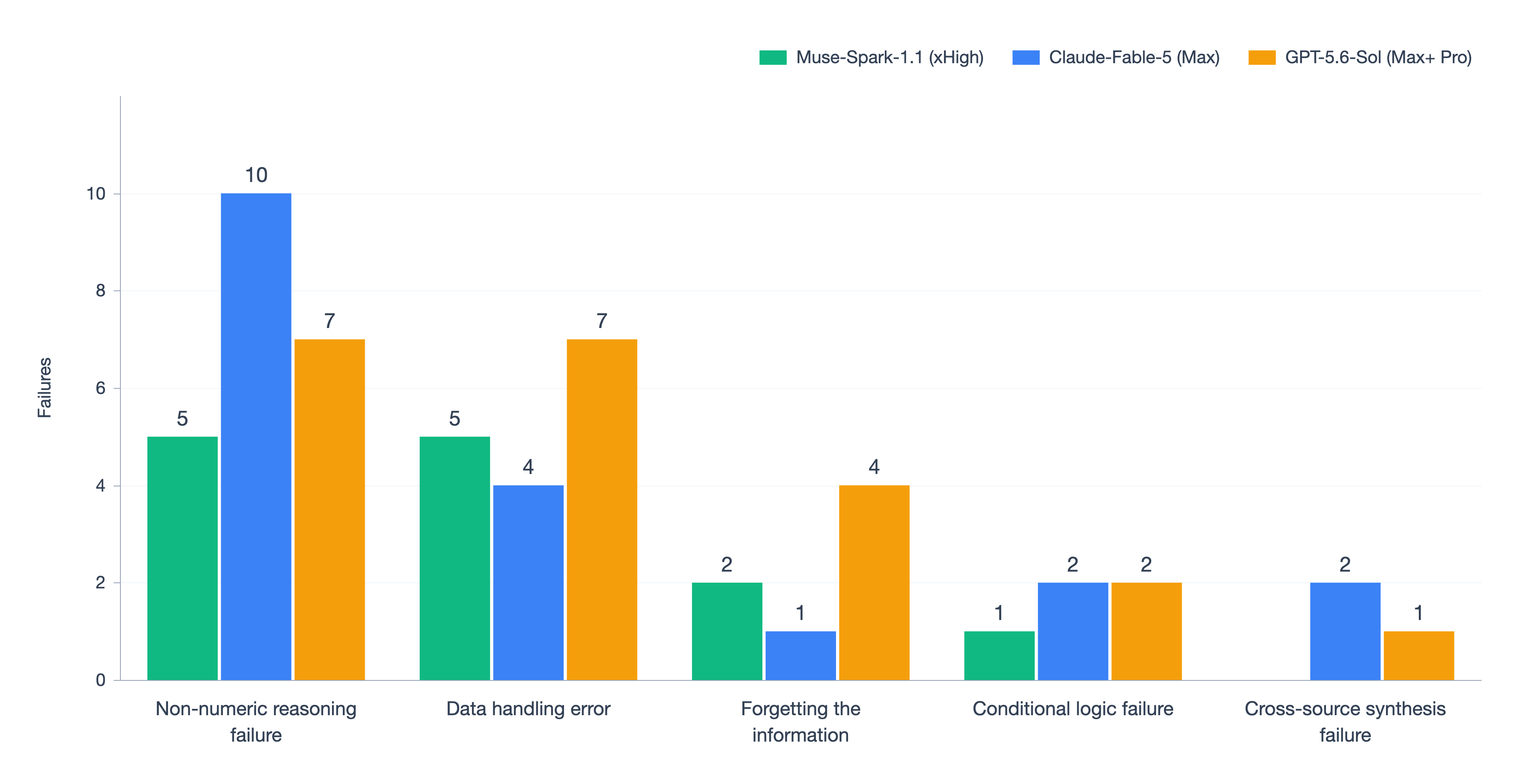}
\label{fig:failure_l2_reasoning}
\end{figure*}

\section{Failure Analysis}
\label{sec:failure_modes}
We developed a taxonomy of failure categories for \apex, working with accounting experts who reviewed low-scoring trajectories.\footnote{The full taxonomy, with definitions for every L1 category and L2 subclass, is in \Cref{app:taxonomy}.} A substantial part of this analysis was spent verifying that failures were real. For each trajectory, experts manually checked the rubric criteria scores to confirm that the model actually erred. Any issues were corrected during production. Once the taxonomy was finalized, we sampled trajectories from the three best-performing models with low mean scores (n=$74$) and categorized each one using an LM judge. \\

\subsection{Failure Modes of the Best-Performing Models}
\label{sec:failure_modes_top_models}
The three best-performing models fail in very similar ways. \Cref{fig:failure_modes_top3} compares their L1 failure profiles over annotated failures ($n=24$ for Claude-Fable-5, $28$ for GPT-5.6-Sol, $22$ for Muse-Spark-1.1). All $74$ failures fall into just four of the taxonomy's seven L1 categories, and reasoning failures dominate every profile: $79\%$ of annotated failures for Claude-Fable-5, $75\%$ for GPT-5.6-Sol, and $59\%$ for Muse-Spark-1.1. Interestingly, not a single annotated failure involves tool use, and information-gathering and instruction-following failures combined account for $17\%$ of Claude-Fable-5's failures and $21\%$ of GPT-5.6-Sol's. Muse-Spark-1.1 is the outlier at $36\%$ combined, the only model whose failures spread meaningfully into incomplete search and dropped requirements. Top models reliably find the right inputs; they make mistakes when applying accounting judgment and multi-step logic to them.\footnote{Full per-model sunburst breakdowns are in \Cref{app:failure_by_top_models}, with exact L2 subclass counts in \Cref{tab:failure_l2_counts}; \Cref{app:taxonomy} defines each label.} This distribution of failure modes likely explains why the \RampHarness{} environment does not yield a significant uplift for these models (\Cref{sec:harness_experiments}). The non-numeric reasoning failures and data handling errors, which account for $51.4\%$ of all annotated failures ($38$ of $74$) are largely driven by fundamental model limitations and not the harness. The harness design primarily addresses environment-based failures (e.g., tool use or context management), which account for at most $17$--$36\%$ of top-model failures.\\

Given that reasoning failures are the modal category, \Cref{fig:failure_l2_reasoning} breaks them down by L2 subclass. Two subclasses dominate: non-numeric reasoning failures ($22$ of the $74$ failures) and data handling errors ($16$ of $74$) jointly account for just over half of all annotated failures. We observe one pattern repeatedly: the model retrieves the correct documents, often computes the correct intermediate values, then substitutes the wrong basis, source, or authorization logic downstream. For example, Claude-Fable-5 treats a program director's report as written donor authorization and releases a restricted $\$15{,}000$ balance; GPT-5.6-Sol shrinks a fixed room-night denominator by out-of-order rooms; Muse-Spark-1.1 re-normalizes allocation weights after computing the correct allocation. Forgetting the information, deriving a correct intermediate result and then dropping or contradicting it downstream, happens in all three profiles ($7$ cases) and is GPT-5.6-Sol's third-largest reasoning failure mode. \\

The skews differ: Claude-Fable-5 has more non-numeric reasoning failures ($53\%$ of all its reasoning failures), whereas GPT-5.6-Sol tilts toward data handling and forgetting failures, and Muse-Spark-1.1 splits evenly between data handling and non-numeric reasoning failures. Some tasks cause the same failures in multiple models: on a tenant-onboarding task, Claude-Fable-5 and GPT-5.6-Sol both report the new lease's square footage as the entity-wide total; and on a WIP-to-finished-goods transfer, GPT-5.6-Sol and Muse-Spark-1.1 both post an incorrect cost variance, taken from the same actual-cost source. \\ 

\subsection{Per-Task Headroom}
\label{sec:gross_headroom}
\begin{figure*}[t]
\centering
\caption{Gross headroom for each model pair: Gross headroom counts tasks where the row model achieved a perfect pass on at least one run but the column model never did; only tasks attempted by the column model are included, partial passes receive no credit, and ties are excluded. Models are ordered by leaderboard rank.}
\includegraphics[width=0.9\textwidth]{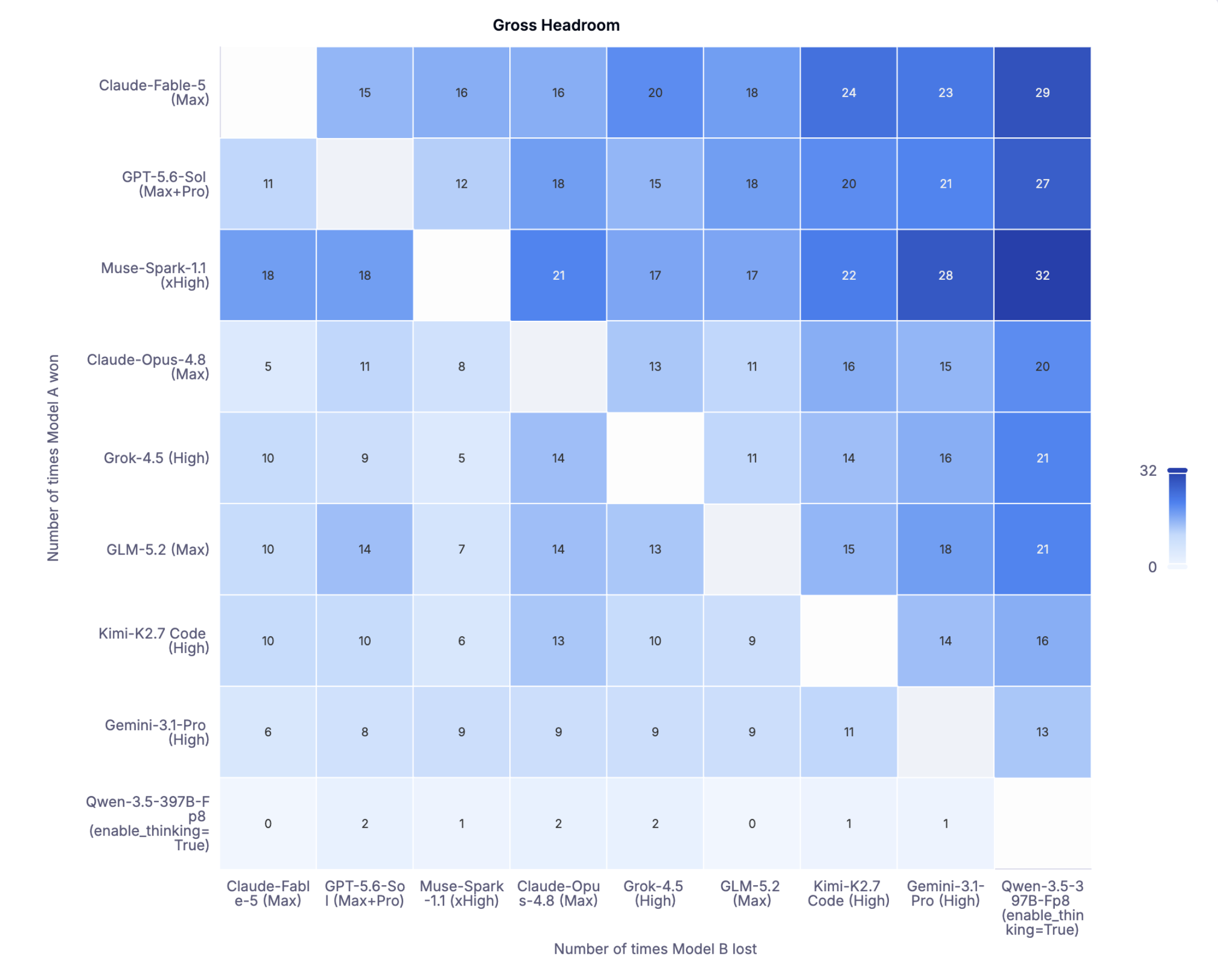}
\label{fig:gross_headroom}
\end{figure*}

We compare models' task win rates head-to-head. For each pair, \Cref{fig:gross_headroom} counts the tasks on which the row model outscores the column model. Because we report gross wins rather than a net difference, the matrix shows headroom in both directions. For example, Claude-Fable-5 outscores Muse-Spark-1.1 on $16$ tasks while Muse-Spark-1.1 wins $18$ back, a tenth of the benchmark going in each direction. On the other $126$ tasks they tie. Wins over the leader extend well down the roster: Grok-4.5, GLM-5.2, and Kimi-K2.7-Code each beat Claude-Fable-5 on $10$ tasks despite trailing it by $13$+ points on mean score. Muse-Spark-1.1 posts the largest single-pair headroom ($32$ tasks over Qwen3.5-397B, a fifth of the benchmark) and wins more tasks than it loses against every other model despite trailing Claude-Fable-5 on mean score: Muse-Spark-1.1 wins more tasks, Claude-Fable-5 wins by larger margins. $93$ of the $\NTasks{}$ tasks (58\%) are not fully solved by any model in any run.

\section{Limitations}
APEX\textendash Accounting has several limitations. Its four task categories (Reconciliation, Data Entry, Variance Analysis, and Schedules \& Accruals) cover close-cycle bookkeeping work; we leave out tax, audit, consolidation, multi-entity and multi-currency work, and external reporting. We also designed tasks to be completed end-to-end and excluded any tasks that require a human-in-the-loop. This removes a skill that real staff accountants exercise constantly; knowing when to ask a clarifying question rather than proceed on an assumption. This would increase realism if addressed in the task design. With only $\NTasks{}$ held-out tasks across $\NWorlds{}$ worlds, statistical power is limited. After Benjamini--Hochberg correction, $34$ of the $\NPairwiseTests{}$ pairwise Mean Criteria@\KSubsample{} differences remain significant, but two adjacent-rank pairs are statistically indistinguishable (\Cref{app:pairwise}), so the exact ordering at those boundaries should not be over-read. \\

Tasks were selected by filtering from a pool of worlds, with tasks already quality controlled for realism and diversity, based on three frontier models (\SecondModel{}, GPT-5.5, and Gemini-3.1-Pro~Preview) achieving low scores when graded (\Cref{app:difficulty_selection}). We adopted this design to ensure that only worlds with challenging tasks are selected for the benchmark. However, it introduces a risk that the models used to filter the tasks, as well as others in the same family, have depressed scores; and the effect does not transfer uniformly to models from other families. We do not believe that the leaderboard ordering is an artifact of the filter: we selected tasks based on filtering over all three models rather than indexing on any single family's idiosyncratic failures. Further, \TopModel{}, the successor of filtering model \SecondModel{}, ranks first, and \ThirdModel{}, the successor of GPT-5.5, ranks third. Thus, being selected against does not by itself determine a family's rank. \\

As with any LM-as-a-judge, grading is subject to judge error. We validated \JudgeModel{} extensively before deployment, benchmarking $\JudgeCandidateCount{}$ candidate judges against $\JudgeLabelN{}$ majority-vote expert labels and optimizing its prompt. It reaches $\JudgeAccuracy{}$ agreement with the expert ground truth ($F_1 = 0.970$), which is very strong but not perfect fidelity. The judge makes both false-positive and false-negative errors (precision = 0.966, recall = 0.974 on the \MetLabel{} class), so absolute scores differ slightly from expected ground truth labels. Because these residual errors are small and are applied uniformly across all evaluated models, they are unlikely to materially affect relative rankings.

\section{Conclusion}
\apex assesses whether frontier models can perform the real work of accountants: finding and reading the right files, applying accounting judgment, and carrying multi-step analyses through to a final close. The best model reaches \TopScore{} Mean Criteria@\KSubsample{}, no model exceeds \MaxPassK{} Pass\textasciicircum{\NRuns{}}, and $58\%$ of tasks are not fully solved by any model in any run. An agent that is usually correct but not reliably so cannot yet close the books unsupervised. \\

Our failure analysis shows that the best-performing models can often retrieve the right inputs; they fail by mishandling multi-step reasoning over those inputs: substituting the wrong basis or authorization logic, and dropping correct intermediate results before the final answer. Tool call errors are not widespread. As such, we anticipate that progress will come primarily from improving the models themselves: accounting-specific training that instills the discipline to carry results through, surface contradictions in the documents, and refuse to post an entry the evidence does not support.

\bibliography{apexcode}

\begin{thebibliography}{15}
\expandafter\ifx\csname natexlab\endcsname\relax\def\natexlab#1{#1}\fi

\bibitem[{Agrawal et~al.(2026)Agrawal, Tan, Soylu, Ziems, Khare, Opsahl-Ong,
  Singhvi, Shandilya, Ryan, Jiang, Potts, Sen, Dimakis, Stoica, Klein, Zaharia,
  and Khattab}]{gepa}
Lakshya~A. Agrawal, Shangyin Tan, Dilara Soylu, Noah Ziems, Rishi Khare, Krista
  Opsahl-Ong, Arnav Singhvi, Herumb Shandilya, Michael~J. Ryan, Meng Jiang,
  Christopher Potts, Koushik Sen, Alexandros~G. Dimakis, Ion Stoica, Dan Klein,
  Matei Zaharia, and Omar Khattab. 2026.
\newblock \href {http://arxiv.org/abs/2507.19457} {{GEPA}: Reflective prompt
  evolution can outperform reinforcement learning}.
\newblock In \emph{International Conference on Learning Representations
  (ICLR)}.

\bibitem[{Azp{\'u}rua(2026)}]{azpurua2026enhance}
Ana~Elena Azp{\'u}rua. 2026.
\newblock \href
  {https://www.library.hbs.edu/working-knowledge/enhance-or-eliminate-how-ai-will-likely-change-these-jobs}
  {Enhance or eliminate? {How} {AI} will likely change these jobs}.
\newblock Harvard Business School Working Knowledge.

\bibitem[{Chen et~al.(2024)Chen, Srinivasan, and
  Zakerinia}]{chen2024displacement}
Wilbur~Xinyuan Chen, Suraj Srinivasan, and Saleh Zakerinia. 2024.
\newblock \href
  {https://www.hbs.edu/ris/Publication%20Files/25-039_05fbec84-1f23-459b-8410-e3cd7ab6c88a.pdf}
  {Displacement or complementarity? {The} labor market impact of generative
  {AI}}.
\newblock Working Paper 25-039, Harvard Business School.

\bibitem[{Choi and Xie(2026)}]{choi2025humanai}
Jung~Ho Choi and Chloe~L. Xie. 2026.
\newblock \href {https://doi.org/10.1111/1475-679X.70052} {Human + {AI} in
  accounting: Early evidence from the field}.
\newblock \emph{Journal of Accounting Research}, 64(3):1333--1373.

\bibitem[{Eulerich et~al.(2024)Eulerich, Sanatizadeh, Vakilzadeh, and
  Wood}]{eulerich2024hype}
Marc Eulerich, Aida Sanatizadeh, Hamid Vakilzadeh, and David~A. Wood. 2024.
\newblock \href {https://doi.org/10.1007/s11142-024-09833-9} {Is it all hype?
  {ChatGPT}'s performance and disruptive potential in the accounting and
  auditing industries}.
\newblock \emph{Review of Accounting Studies}, 29(3):2318--2349.

\bibitem[{Goyal(2025)}]{goyal2025whileloop}
Ankur Goyal. 2025.
\newblock \href {https://www.braintrust.dev/blog/agent-while-loop} {The
  canonical agent architecture: A while loop with tools}.
\newblock Braintrust Blog.

\bibitem[{Hughes(2024)}]{hughes2024takeaway}
Patrick Hughes. 2024.
\newblock \href
  {https://hl.com/insights/the-takeaway-a-qa-with-patrick-hughes-on-private-equity-s-rising-interest-in-accounting-firms/}
  {The takeaway: A {Q\&A} with {Patrick Hughes} on private equity's rising
  interest in accounting firms}.
\newblock Houlihan Lokey Insights.

\bibitem[{Jiang et~al.(2025)Jiang, Yang, Cui, Jin, Wang, Li, Huang, Sun, and
  Wang}]{jiang2025finmaster}
Junzhe Jiang, Chang Yang, Aixin Cui, Sihan Jin, Ruiyu Wang, Bo~Li, Xiao Huang,
  Dongning Sun, and Xinrun Wang. 2025.
\newblock \href {http://arxiv.org/abs/2505.13533} {Finmaster: A holistic
  benchmark for mastering full-pipeline financial workflows with llms}.

\bibitem[{{National Center for O*NET Development}(2025)}]{onet2025accountants}
{National Center for O*NET Development}. 2025.
\newblock \href {https://www.onetonline.org/link/summary/13-2011.00}
  {13-2011.00 --- accountants and auditors}.
\newblock O*NET Online.

\bibitem[{Patwardhan et~al.(2025)Patwardhan, Dias, Proehl, Kim, Wang, Watkins,
  Fishman, Aljubeh, Thacker, Fauconnet, Kim, Chao, Miserendino, Chabot, Li,
  Sharman, Barr, Glaese, and Tworek}]{patwardhan2025gdpval}
Tejal Patwardhan, Rachel Dias, Elizabeth Proehl, Grace Kim, Michele Wang,
  Olivia Watkins, Sim{\'o}n~Posada Fishman, Marwan Aljubeh, Phoebe Thacker,
  Laurance Fauconnet, Natalie~S. Kim, Patrick Chao, Samuel Miserendino, Gildas
  Chabot, David Li, Michael Sharman, Alexandra Barr, Amelia Glaese, and Jerry
  Tworek. 2025.
\newblock \href {http://arxiv.org/abs/2510.04374} {{GDPval}: Evaluating ai
  model performance on real-world economically valuable tasks}.

\bibitem[{{Penrose}(2025)}]{penrose2025accountingbench}
{Penrose}. 2025.
\newblock \href {https://accounting.penrose.com/} {{AccountingBench}:
  Evaluating {LLMs} on real long-horizon business tasks}.
\newblock Blog post.

\bibitem[{Stevens(2026)}]{stevens2026stackbenchmarking}
Ryan Stevens. 2026.
\newblock \href {https://builders.ramp.com/post/stack-benchmarking} {Stack
  benchmarking}.
\newblock Ramp Builders.

\bibitem[{Wang et~al.(2025)Wang, Liu, Zhao, Li, and Zhang}]{wang2025auditbench}
Rushi Wang, Jiateng Liu, Weijie Zhao, Shenglan Li, and Denghui Zhang. 2025.
\newblock \href {https://doi.org/10.1007/978-981-96-8912-5_3} {{AuditBench}: A
  benchmark for large language models in financial statement auditing}.
\newblock In \emph{AI for Research and Scalable, Efficient Systems (AI4Research
  and SEAS Workshops at AAAI 2025)}, volume 2533 of \emph{Communications in
  Computer and Information Science}, pages 59--81. Springer Nature Singapore.
\newblock Also available as arXiv:2506.17282 under the title ``Automating
  Financial Statement Audits with Large Language Models''.

\bibitem[{Yao et~al.(2025)Yao, Shinn, Razavi, and Narasimhan}]{yao2024tau}
Shunyu Yao, Noah Shinn, Pedram Razavi, and Karthik Narasimhan. 2025.
\newblock {$\tau$-bench}: A benchmark for tool-agent-user interaction in
  real-world domains.
\newblock In \emph{International Conference on Learning Representations
  (ICLR)}.

\bibitem[{Yao et~al.(2023)Yao, Zhao, Yu, Du, Shafran, Narasimhan, and
  Cao}]{yao2023react}
Shunyu Yao, Jeffrey Zhao, Dian Yu, Nan Du, Izhak Shafran, Karthik Narasimhan,
  and Yuan Cao. 2023.
\newblock \href {https://arxiv.org/abs/2210.03629} {{ReAct}: Synergizing
  reasoning and acting in language models}.
\newblock In \emph{International Conference on Learning Representations
  (ICLR)}.

\end{thebibliography}
\bibliographystyle{acl_natbib}
\FloatBarrier
\clearpage

\appendix

\section{Worked Task Example}
\label{app:worked_example}

To make task and rubric structure concrete, we walk through one dev set task
end to end: the prompt as given to the agent, two representative rubric
criteria, and a condensed trajectory from one evaluated model. The example
illustrates how binary, outcome-based criteria grade an open-ended
deliverable.

\subsection{Example Prompt}
\label{app:worked_example_prompt}

The prompt below is the complete task input: the agent receives no file
list and must locate its own evidence in the world's file system.

\begin{tcolorbox}[breakable, colback=gray!5, colframe=black!60,
  title=Task: \texttt{World 9\_Task 14}, fonttitle=\bfseries]
\textbf{Category:} Reconciliation \\

\textit{Review the December client cost advance reconciliation.
Reconcile the client cost advance ledger to the QBO/GL balance for client
cost advances. Using the cost advance ledger methodology and provided
support, verify whether hard costs that should be treated as client cost
advances were capitalized and whether soft costs were expensed. Identify any
discrepancies in cost classification or capitalization treatment that need
to be resolved.}\\

\textit{Round all amounts to two decimal places. Do not round intermediary
values. Provide your answer to the console.}
\end{tcolorbox}

\subsection{Example Rubric Criteria}
\label{app:worked_example_rubric}

\Cref{tab:worked_example_rubric} shows two of the $24$ binary criteria used
to grade this task: one qualitative criterion that grades whether the
response identifies the task's central seeded error (contract attorney
fees expensed rather than capitalized), and a numerical criterion that
grades a supporting dollar amount for the same finding. The trajectory
excerpted in \Cref{app:worked_example_trajectory} fails both.

\begin{table}[h]
\centering
\small
\caption{Two representative rubric criteria for the worked example task (of
$24$ total).}
\label{tab:worked_example_rubric}
\begin{tabularx}{\linewidth}{@{}lX@{}}
\toprule
\textbf{Type} & \textbf{Criterion} \\
\midrule
Qualitative & Identifies the Whitfield M-2024-0018 Contract Attorney Fees
Matter Direct (or Account 5000) AP bill lines as a capitalization issue. \\
\addlinespace
Numerical & States that the Whitfield M-2024-0018 account 5000 contract attorney matter direct lines total \$23,500.00 (acceptable value is \$23,500.00). \\
\bottomrule
\end{tabularx}
\end{table}

\subsection{Example Trajectory}
\label{app:worked_example_trajectory}

The excerpt below shows six representative turns from one evaluated 
trajectory (Muse-Spark-1.1, \LoopHarness{}). It omits the remaining turns, 
tool calls, and intervening reasoning. The full trajectory comprises
$19$ turns containing $41$ tool calls; under this harness, an assistant
message with no tool calls is treated as the final answer and ends the
run. The model met $11$ of the task's $24$ criteria ($45.8\%$).

\begin{tcolorbox}[breakable, colback=gray!5, colframe=black!60,
  title={Trajectory excerpt --- Muse-Spark-1.1, \LoopHarness{}, $11$/$24$ criteria met}]
\small\raggedright
\textbf{Turn 1.} \textit{``I've mapped the client cost advance task. I'll
start by locating your reconciliation files and ledger.''} \newline
\texttt{Tool: \seqsplit{filesystem\_list\_files(/)}}

\vspace{0.4em}
\textbf{Turn 3.} \textit{``I've located the reconciliation workpapers. Now
I'll dig into the cost advance methodology and GL balances to uncover the
discrepancies.''} \newline
\texttt{Tool: \seqsplit{excel\_read\_tab(workpaper\_client\_cost\_advance\_ledger\_2024.xlsx)} $\times 5$ tabs}

\vspace{0.4em}
\textbf{Turn 4.} \textit{``I've mapped the cost advance methodology ---
trust-paid vs.\ firm-advanced and hard vs.\ soft. Now I'll reconcile the
ledger to GL and verify capitalization treatment.''} \newline
\texttt{Tools: \seqsplit{excel\_read\_tab(qbo\_closing\_trial\_balance\_2024\_12\_31.xlsx)},
\seqsplit{excel\_read\_tab(workpaper\_whitfield\_settlement\_summary.xlsx)},
\seqsplit{filesystem\_list\_files(/contract\_attorney\_invoices)}}

\vspace{0.4em}
\textbf{Turn 8.} \textit{``I've traced the Whitfield settlement and
trust-funded costs. Now I'll verify contingency matter capitalization and
hard vs.\ soft cost treatment.''} \newline
\texttt{Tool: \seqsplit{excel\_filter\_tab(qbo\_general\_ledger\_detail\_2024.xlsx,} Matter contains \seqsplit{M-2024-0018)} $\times 2$}

\vspace{0.4em}
\textbf{Turn 13} (reasoning written as comments in a code cell)\textbf{.}
\textit{``In our earlier search we only found Whitfield bills (Westfield
Legal Services, Jennifer Kao, Marcus Avila). But what about other
contingency matters? Meadowbrook and Ferndale have `matter file maintained
off-system' --- perhaps there are costs in AP bills that should be
capitalized to 1300 for those matters but were expensed?''} \newline
\texttt{Tool: \seqsplit{code\_execution\_run\_shell(python3:} scan GL detail for AP bills on contingency matters\texttt{)}}

\vspace{0.4em}
\textbf{Turn 19 (final answer -- a message with no tool calls).}
\textit{``QBO/GL balance 1300 = \$0.00 reconciles to Client Cost Advance
Ledger = \$0.00. Tie-outs to IOLTA (\$29,400.00) and Clio (\$47,587.09) and
settlement (\$45,000.00) pass. Only classification correction required is
re-labeling three Heritage transactions (\$2,950.00) from Soft to Hard in
ledger workpaper. No dollar adjustment to GL required at Dec 31, 2024,
pending review of off-system Meadowbrook/Ferndale registers.''}
\end{tcolorbox}

The model ties the ledger to the GL, verifies the supporting tie-outs, and
catches a minor workpaper labeling issue (\$2,950.00), earning $11$ of
$24$ criteria. But it misses the error the task is built around:
\$23,500.00 of Whitfield contract attorney fees expensed to account 5000
instead of capitalized to account 1300. \\

The model retrieved the relevant bills and ledger records but did not open the cost-recovery addendum containing the controlling authorization. We therefore classify the primary failure as incomplete information gathering. A secondary reasoning failure followed: the model treated the zero account-1300 balance as evidence that no capitalization adjustment was required, rather than investigating whether relevant costs had been misclassified.

\section{Quality Control}
\label{app:quality_control}
\begin{figure*}[t]
\centering
\caption{Our world construction and audit process consisted of expert and synthetic generation with multiple rounds of expert audits.}
\includegraphics[width=\textwidth]{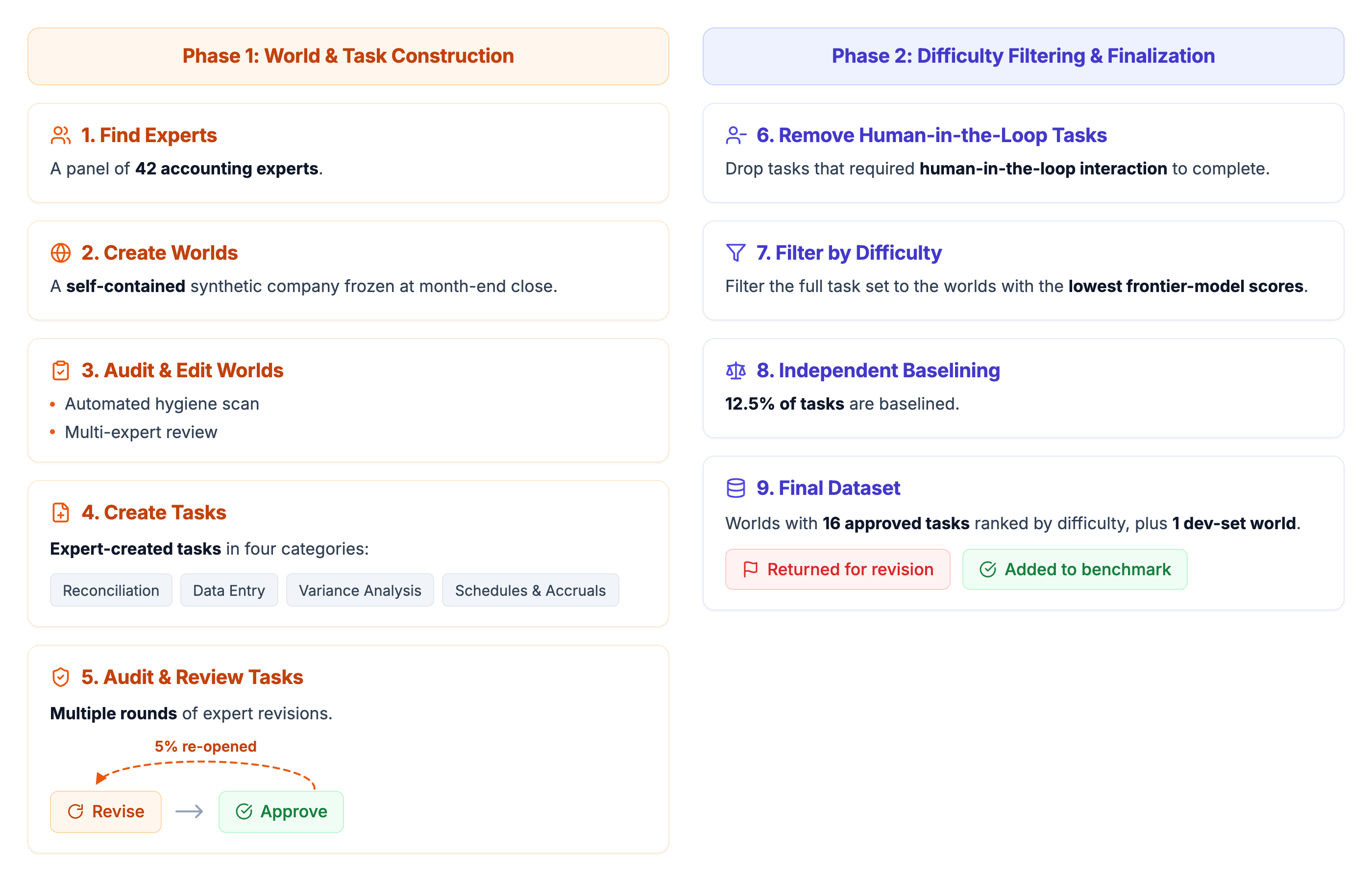}
\label{fig:pipeline}
\end{figure*}

This appendix section details the quality-control process behind \Cref{sec:benchmark_construction}. The aim was to verify that every task is solvable from the shipped world files alone and admits a single defensible answer, so that rubric grades measure model ability rather than dataset defects. Quality control was run at both the world and task level (\Cref{fig:pipeline}). At the world level, four checks were applied:

\begin{enumerate}
  \item An automated hygiene pass scanned for blatant contradictions and data-hygiene problems.
  \item Multiple experts manually reviewed all files for realism and consistency.
  \item A validation script confirmed the accounting software data was schema-compatible with the evaluation environment.
  \item Three to five experts each completed a handful of tasks against the world files, surfacing contradictions and missing data by having professionals actually do the work.
\end{enumerate}

At the task level, each submission passed through one layer of expert review (with revision cycles as needed) and an automated QC pass. Tasks failing automated QC were audited by the project team, and roughly $\TaskReauditRate{}$ of approved tasks were manually re-reviewed throughout the project lifecycle. \\

As a final check, we baselined $20$ tasks with independent experts who had no prior exposure to the task, giving them only the prompt and world files and having them re-solve the task from scratch. Comparing each independent answer against the shipped golden response surfaced a single divergent task, on which the independent re-solve disagreed with the shipped golden on $3$ of its $19$ rubric criteria ($5\%$ defect rate by task). Across all $20$ baselined tasks this was the only disagreement observed, an error rate of $3$ of $276$ total criteria ($1.1\%$ defect rate by criteria). The divergent task was returned for revision. This check confirmed that the remaining baselined tasks were fully solvable from the shipped files alone and admitted a single defensible answer.

\subsection{Difficulty and Selection}
\label{app:difficulty_selection}
Prior to baselining, the final benchmark was filtered from a larger authored pool of $\CandidatePoolSize{}$ candidate tasks, on which frontier models scored consistently high under the \LoopHarness{}, leaving little headroom to separate them, so we retained only the hardest worlds. We restricted the benchmark to worlds with $\TasksPerWorld{}$ or more approved tasks and selected the $\QualifyingWorlds{}$ with the lowest average Mean Criteria@\KSubsample{} score among their $\TasksPerWorld{}$ hardest tasks, using \SecondModel{}, GPT-5.5, and Gemini-3.1-Pro~Preview as the filtering models; the $\NWorlds{}$ most difficult worlds formed the benchmark, and the $\QualifyingWorlds{}$th (a legal-services world) became the public dev set, from which we release $\DevSetTasks{}$ tasks.\\

Before filtering for difficulty, we removed $\IncompleteTasksRemoved{}$ incomplete-information tasks, those where a model is meant to refuse service or ask for clarification, given their small representation in the dataset and the human-in-the-loop support they would require.

\section{World and Task Statistics}
\label{app:world_stats}
This appendix reports per-world scale statistics (\Cref{tab:world_task_criteria_stats}). We checked whether task complexity (e.g., criteria counts and required files) was spread evenly across worlds, so that no single world dominates the benchmark. \\

\begin{table*}[t]
\raggedright
\scriptsize
\setlength{\tabcolsep}{4pt}
\caption{Task, criteria, and file-type statistics by world.}
\label{tab:world_task_criteria_stats}
\resizebox{\textwidth}{!}{%
\begin{tabular}{l|cccccccc}
\toprule
\textbf{World}
& \textbf{Tasks}
& \makecell{\textbf{Total}\\\textbf{Criteria}}
& \makecell{\textbf{Unique}\\\textbf{Files}}
& \makecell{\textbf{Unique}\\\textbf{Spreadsheets}}
& \makecell{\textbf{Unique}\\\textbf{PDFs}}
& \makecell{\textbf{Unique}\\\textbf{QuickBooks}}
& \makecell{\textbf{Mean}\\\textbf{Criteria/Task}}
& \makecell{\textbf{Mean}\\\textbf{Files/Task}} \\
\midrule
\rowcolor{gray!15}
World 9 (dev set)  & $10$ & $89$  & $90$  & $34$ & $46$  & $10$  & $8.9$  & $5.6$  \\
World 2.2  & $16$ & $183$ & $64$  & $33$ & $27$ & $2$  & $11.4$ & $6.6$  \\
World 3    & $16$ & $297$ & $79$  & $43$ & $27$ & $5$  & $18.6$ & $6.2$  \\
World 5.2  & $16$ & $190$ & $61$  & $27$ & $23$ & $8$  & $11.9$ & $6.0$  \\
World 6.2  & $16$ & $237$ & $64$  & $40$ & $18$ & $5$  & $14.8$ & $7.0$  \\
World 7    & $16$ & $317$ & $166$ & $85$ & $71$ & $0$  & $19.8$ & $14.1$ \\
World 8    & $16$ & $186$ & $65$  & $30$ & $24$ & $10$ & $11.6$ & $7.6$  \\
World 10   & $16$ & $180$ & $54$  & $28$ & $17$ & $7$  & $11.3$ & $6.0$  \\
World 11   & $16$ & $212$ & $72$  & $32$ & $36$ & $0$  & $13.3$ & $8.4$  \\
World 13   & $16$ & $191$ & $54$  & $30$ & $17$ & $5$  & $11.9$ & $6.7$  \\
World 14   & $16$ & $193$ & $52$  & $32$ & $16$ & $3$  & $12.1$ & $6.5$  \\
\midrule
\textbf{Mean (w/ dev set)}    & $-$ & $207$ & $74.6$ & $37.6$ & $29.3$ & $5.0$ & $13.4$ & $7.4$ \\
\textbf{Median (w/ dev set)}  & $-$ & $191$ & $64$    & $32$    & $24$    & $5$    & $11.9$    & $6.6$    \\
\midrule
\textbf{Mean (w/o dev set)}   & $-$ & $219$ & $73.1$ & $38.0$ & $27.6$ & $4.5$ & $13.7$ & $7.5$ \\
\textbf{Median (w/o dev set)} & $-$ & $192$ & $64$ & $32$ & $23.5$  & $5$    & $12$    & $6.7$    \\
\bottomrule
\end{tabular}%
}
\end{table*}

\apex{} tasks fall into four categories (\Cref{tab:dataset_stats}); \Cref{tab:devset_category_stats} reports the same breakdown restricted to the public dev set world (World 9), which has a smaller and less evenly distributed sample per category ($n=\DevSetTasks{}$ total) than the held-out benchmark. World 9 simulates a boutique law firm completing its December month-end close under U.S. GAAP accrual accounting; its tasks cover law-firm-specific close work such as trust (IOLTA) accounting, client cost advances, and partner-compensation accruals.

\begin{table}[t]
\centering
\scriptsize
\caption{\apex{} dev set (World 9) tasks by category ($n=\DevSetTasks{}$).}
\label{tab:devset_category_stats}
\setlength{\tabcolsep}{3pt}
\renewcommand{\arraystretch}{1.02}
\resizebox{\columnwidth}{!}{%
\begin{tabular}{lccccc}
\toprule
\textbf{Category} & \textbf{Tasks} & \makecell{\textbf{Mean}\\\textbf{Criteria}} & \makecell{\textbf{Median}\\\textbf{Criteria}} & \makecell{\textbf{Mean}\\\textbf{Files}} & \makecell{\textbf{Median}\\\textbf{Files}} \\
\midrule
Data Entry             & $1$ & $2$    & $2$    & $5$   & $5$   \\
Reconciliation         & $4$ & $11.5$ & $9$    & $7$   & $7.5$ \\
Variance Analysis      & $2$ & $10.5$ & $10.5$ & $2.5$ & $2.5$ \\
Schedules \& Accruals  & $3$ & $6.7$  & $5$    & $6$   & $6$   \\
\midrule
Overall                & $\DevSetTasks{}$ & $8.9$ & $5.5$ & $5.6$ & $5.5$ \\
\bottomrule
\end{tabular}%
}
\end{table}

\section{Model Configurations}
\label{app:model_details}
This appendix section records the exact model configurations, context limits, and thinking-effort settings used for the leaderboard runs, for reproducibility. All models are called via LiteLLM with retry logic (exponential backoff) and a maximum of three attempts per request. Model configurations are described in Table~\ref{tab:model_details}.

\begin{table*}[t]
\centering
\small
\caption{Information about the models evaluated on the APEX\textendash Accounting leaderboard: provider, context window, maximum output, thinking configuration, and mean/median wall-clock time per run.}
\label{tab:model_details}
\setlength{\tabcolsep}{6pt}
\renewcommand{\arraystretch}{1.15}
\begin{tabular}{l|c|ccc>{\raggedright\arraybackslash}p{4.2cm}|cc}
\toprule
\textbf{Model} & \textbf{Provider} & \makecell{\textbf{Context}\\\textbf{Window}} & \makecell{\textbf{Max}\\\textbf{Output}} & \textbf{Temp.} & \textbf{Thinking} & \makecell{\textbf{Mean}\\\textbf{min/run}} & \makecell{\textbf{Median}\\\textbf{min/run}} \\
\midrule
\TopModel{}        & Anthropic  & 1M   & 128k & Default & adaptive, effort=max            & $22.0$ & $20.5$ \\
Muse-Spark-1.1     & Meta       & ---  & ---  & $1.0$ & reasoning\_effort=xHigh          & $80.9$ & $76.1$ \\
\ThirdModel{}      & OpenAI     & 1M   & ---  & $1.0$ & reasoning mode=pro, effort=max  & $54.3$ & $46.6$ \\
\SecondModel{}     & Anthropic  & 1M   & 128k & Default & adaptive, effort=max            & $44.3$ & $34.9$ \\
GLM-5.2            & Z.AI  & 1M   & 131k & $1.0$ & reasoning\_effort=max            & $43.4$ & $32.9$ \\
Grok-4.5           & xAI        & 500k & ---  & $1.0$ & default (high)         & $12.0$ & $9.4$  \\
Kimi-K2.7-Code     & MoonshotAI  & 256k & ---  & $1.0$ & reasoning\_effort=high           & $11.6$ & $10.2$ \\
Gemini-3.1-Pro     & Google  & 1M   & 32k  & $1.0$ & reasoning\_effort=high (+thoughts) & $28.4$ & $29.9$ \\
Qwen3.5-397B-FP8       & Alibaba      & ---  & 82k  & $0.6$ & enable\_thinking=True            & $60.5$ & $43.4$ \\
\bottomrule
\end{tabular}%
\end{table*}

\section{Judge Evaluation Details}
\label{app:judge_eval}
The question behind this comparison is whether a low-cost model (not on the leaderboard roster) can match frontier-model grading agreement. \Cref{tab:llmaaj_scores} reports full comparison results across all $\JudgeCandidateCount{}$ candidate judges with a GEPA-optimized~\citep{gepa} grading prompt evaluated in \Cref{sec:methodology}, scored against majority-vote human ground truth. \JudgeModel{} (the deployed judge, shown in bold) matches the agreement of substantially more expensive proprietary judges---its $F_1$ of $0.970$ matches Claude-Opus-4.8 with high reasoning and exceeds every other individual judge tested.

\begin{table}[t]
\centering
\scriptsize
\setlength{\tabcolsep}{3pt}
\renewcommand{\arraystretch}{1.02}
\caption{Full LM-as-a-judge comparison against the majority-vote human ground truth on the \MetLabel{} class. Every judge in this comparison was evaluated with the same GEPA-optimized~\citep{gepa} grading prompt. The deployed judge, \JudgeModel{}, is shown in bold. The last three rows are ensemble configurations which combine several judges' verdicts by majority vote: Open Source aggregates the open-weight judges, Oracle Models the closed frontier judges, and Majority Vote all judges together. (High/Low denote reasoning effort settings.)}

\label{tab:llmaaj_scores}
\resizebox{\columnwidth}{!}{%
\begin{tabular}{lcccc}
\toprule
\textbf{Judge} & \textbf{Prec.} & \textbf{Rec.} & \textbf{$F_1$} & \textbf{Acc.} \\
\midrule
\textbf{DeepSeek-v4-Flash (High)}  & $\mathbf{0.966}$ & $\mathbf{0.974}$ & $\mathbf{0.970}$ & $\mathbf{0.971}$ \\
Claude-Opus-4.8 (High)      & $0.946$ & $0.994$ & $0.970$ & $0.970$ \\
Claude-Opus-4.8 (Low)       & $0.947$ & $0.993$ & $0.969$ & $0.969$ \\
Gemini-3.1-Pro~(High)       & $0.945$ & $0.994$ & $0.969$ & $0.969$ \\
Gemini-3.1-Pro~(Low)        & $0.946$ & $0.989$ & $0.967$ & $0.967$ \\
DeepSeek-v4-Pro (High)      & $0.965$ & $0.968$ & $0.967$ & $0.967$ \\
Kimi-K2.6                   & $0.973$ & $0.960$ & $0.966$ & $0.967$ \\
Gemini-3-Flash              & $0.938$ & $0.994$ & $0.965$ & $0.965$ \\
GPT-5.5~(High)              & $0.950$ & $0.977$ & $0.963$ & $0.964$ \\
GPT-5.5~(Low)               & $0.948$ & $0.974$ & $0.961$ & $0.962$ \\
\midrule
Open Source (ensemble)      & $0.975$ & $0.985$ & $0.980$ & $0.981$ \\
Oracle Models (ensemble)    & $0.947$ & $0.991$ & $0.968$ & $0.969$ \\
Majority Vote (ensemble)    & $0.954$ & $0.994$ & $0.974$ & $0.974$ \\
\bottomrule
\end{tabular}%
}
\end{table}

\subsection{Judge Agreement by Solver Model}
\label{app:judge_by_solver}
\JudgeModel{} is not among the solver models used in this benchmark's leaderboard (\Cref{sec:methodology}), so self-preference is not a structural concern. Nonetheless, \Cref{tab:judge_by_solver} breaks out judge agreement, precision, and recall separately for each solver model's trajectories as a robustness check.

\begin{table}[t]
\centering
\small
 \caption{\JudgeModel{} agreement against majority-vote ground truth, broken out by solver model.}
\label{tab:judge_by_solver}
\begin{tabular}{lccc}
\toprule
\textbf{Solver Model} & \textbf{Agreement} & \textbf{Precision} & \textbf{Recall} \\
\midrule
\TopModel{}        & 0.980 & 0.984 & 0.981 \\
\ThirdModel{}      & 0.947 & 0.918 & 0.965 \\
Gemini-3.1-Pro     & 0.984 & 0.989 & 0.975 \\
\midrule
Overall            & 0.971 & 0.966 & 0.974 \\
\bottomrule
\end{tabular}
\end{table}

\subsection{Inter-Rater Reliability}
To validate the ground truth, \Cref{tab:interrater} reports inter-rater reliability among the $\JudgeAnnotators{}$ expert annotators before majority-vote aggregation; Fleiss' $\kappa$ of $0.857$ indicates strong agreement beyond chance for two-category ratings.

\subsection{Self-Preference Stress Test}
\label{app:judge_self_solver}
Although \JudgeModel{} is not among the solver models used for the benchmark, we ran an adversarial check to measure self-preference. We ran \JudgeModel{} as a solver on the full \apex{} benchmark and graded its trajectories with the same \JudgeModel{} judge, temperature = $0.1$, and high reasoning.

\JudgeModel{} attains a Mean Criteria@\KSubsample{} of $37.1\%$---essentially matching Kimi-K2.7-Code ($37.0\%$, seventh on the leaderboard) and far below the leader (Claude-Fable-5, $56.4\%$). A
self-favoring judge would likely push its own solver toward the top of the roster; instead it
lands in the lower-middle, consistent with no material self-preference.

\begin{table}[t]
\centering
\small
\renewcommand{\arraystretch}{1.15}
\caption{Inter-rater reliability of the $\JudgeAnnotators{}$ expert annotators on the judge-selection sample, prior to majority-vote aggregation.}
\label{tab:interrater}
\begin{tabularx}{\columnwidth}{@{}Xr@{}}
\toprule
\textbf{Metric} & \textbf{Value} \\
\midrule
Criteria labeled ($\JudgeAnnotators{}$ graders) & $\JudgeCriteriaN$ \\
Unanimous criteria (all $3$ agree) & $1{,}505$ ($89.2\%$) \\
Non-unanimous criteria & $182$ ($10.8\%$) \\
Mean pairwise agreement & $92.8\%$ \\
Fleiss' $\kappa$ ($3$ raters, $2$ categories) & $0.857$ \\
\bottomrule
\end{tabularx}
\end{table}

\section{Run Completion and Scoring of Incomplete Runs}
\label{app:run_completion}
Runs that fail because of the model (e.g., no submission, a malformed or empty answer, or the step or token budget exhausted without submitting) are scored as zero on \emph{every} criterion, so the run contributes $0\%$ to Mean Criteria@\KSubsample{} and is included both the numerator and the denominator. Runs that fail for reasons outside the model's control, such as a provider API error or a system failure, are excluded from every metric, contributing to neither the numerator nor the denominator. Because those runs are dropped rather than scored, a few tasks have fewer than $\NRuns{}$ scored runs for a given model, which is why paired comparisons span $156$--$160$ tasks per pair (\Cref{app:pairwise}) rather than all $\NTasks{}$.

\section{Pairwise Significance Matrix}
\label{app:pairwise}

This appendix section evaluates which leaderboard gaps are statistically reliable once all $\NPairwiseTests{}$ pairwise comparisons are corrected for multiple testing. For each model pair, we compute the paired per-task difference in Mean Criteria@\KSubsample{} over the tasks with complete runs for both models ($156$--$160$ tasks per pair) and correct the resulting $p$-values with the Benjamini--Hochberg procedure at the $\FDRLevel{}$ false-discovery rate. \Cref{tab:pairwise_matrix} reports the full matrix. $34$ of the $\NPairwiseTests{}$ gaps remain significant after correction. The two exceptions are both adjacent-rank pairs: Muse-Spark-1.1 vs.\ \ThirdModel{} ($+1.2$\,pp, CI $[-2.9, 5.1]$) and GLM-5.2 vs.\ Grok-4.5 ($+1.9$\,pp, CI $[-1.6, 5.4]$). Three further adjacent-rank gaps survive only narrowly ($0.01 \leq q < 0.05$): \TopModel{} vs.\ Muse-Spark-1.1 ($+3.8$\,pp), \ThirdModel{} vs.\ \SecondModel{} ($+3.7$\,pp), and Grok-4.5 vs.\ Kimi-K2.7-Code ($+3.8$\,pp). Every comparison spanning two or more leaderboard ranks is significant at $q < 0.01$, so the ordering is reliable everywhere except at these adjacent boundaries.

\begin{table*}[t]
\centering
\small
\setlength{\tabcolsep}{5pt}
\renewcommand{\arraystretch}{1.15}
\caption{Pairwise Mean Criteria@\KSubsample{} differences (row minus column, percentage points, paired per task over the $156$--$160$ tasks with complete runs for both models). Stars are Benjamini--Hochberg-corrected across all $\NPairwiseTests{}$ comparisons: $^{*}\,q<0.05$, $^{**}\,q<0.01$, $^{***}\,q<0.001$; ns $=$ not significant at the $\FDRLevel{}$ false-discovery rate. Models in leaderboard order.}
\label{tab:pairwise_matrix}
\resizebox{\textwidth}{!}{%
\begin{tabular}{l|cccccccc}
\toprule
 & \textbf{Muse-Spark-1.1} & \textbf{GPT-5.6-Sol} & \textbf{Claude-Opus-4.8} & \textbf{GLM-5.2} & \textbf{Grok-4.5} & \textbf{Kimi-K2.7-Code} & \textbf{Gemini-3.1-Pro} & \textbf{Qwen3.5-397B} \\
\midrule
Claude-Fable-5         & $+3.8^{*}$ & $+5.0^{**}$   & $+8.6^{***}$ & $+13.7^{***}$ & $+15.6^{***}$ & $+19.4^{***}$ & $+24.1^{***}$ & $+32.4^{***}$ \\
Muse-Spark-1.1  &            & $+1.2$\,(ns)  & $+5.0^{**}$  & $+9.9^{***}$  & $+11.8^{***}$ & $+15.6^{***}$ & $+20.3^{***}$ & $+28.6^{***}$ \\
GPT-5.6-Sol     &            &               & $+3.7^{*}$   & $+8.8^{***}$  & $+10.7^{***}$ & $+14.5^{***}$ & $+19.1^{***}$ & $+27.4^{***}$ \\
Claude-Opus-4.8 &            &               &              & $+5.0^{**}$   & $+6.9^{***}$  & $+10.7^{***}$ & $+15.4^{***}$ & $+23.6^{***}$ \\
GLM-5.2         &            &               &              &               & $+1.9$\,(ns)  & $+5.7^{***}$  & $+10.4^{***}$ & $+18.5^{***}$ \\
Grok-4.5        &            &               &              &               &               & $+3.8^{*}$    & $+8.4^{***}$  & $+16.6^{***}$ \\
Kimi-K2.7-Code  &            &               &              &               &               &               & $+4.6^{**}$   & $+12.8^{***}$ \\
Gemini-3.1-Pro  &            &               &              &               &               &               &               & $+8.1^{***}$ \\
\bottomrule
\end{tabular}%
}
\end{table*}

\section{Full Leaderboard and Dev Set Shift}
\label{app:devset_shift}
\Cref{tab:leaderboard} reports the full APEX\textendash Accounting leaderboard: Mean Criteria@\KSubsample{}, Pass@1, Pass@\NRuns{}, and Pass\textasciicircum{\NRuns{}} for every model, with $95\%$ bootstrap confidence intervals. \Cref{tab:leaderboard_devset_shift} then compares each model's held-out leaderboard performance against the public dev set world, to surface overfitting risk on the released split. Because the dev set world is, by construction, the easiest of the \QualifyingWorlds{} qualifying worlds - it ranked last under the
difficulty filter (\Cref{app:quality_control}) - scores are expected to run somewhat higher on the dev set.

\begin{table}[t]
\centering
\scriptsize
\setlength{\tabcolsep}{3pt}
\renewcommand{\arraystretch}{1.02}
\caption{Full APEX\textendash Accounting leaderboard: Mean Criteria@\KSubsample{}, Pass@1, Pass@\NRuns{}, and Pass\textasciicircum{\NRuns{}} (\%) for every model, with $95\%$ task-bootstrap confidence intervals. Best value in each column in bold.}
\label{tab:leaderboard}
\resizebox{\columnwidth}{!}{%
\begin{tabular}{lcccc}
\toprule
\textbf{Model} & \textbf{Mean Criteria@\KSubsample{}} & \textbf{Pass@1} & \textbf{Pass@\NRuns{}} & \textbf{Pass\textasciicircum{\NRuns{}}} \\
\midrule
Claude-Fable-5         & $\mathbf{56.4}$ [$52.7$--$60.2$] & $\mathbf{9.5}$ [$6.0$--$13.4$] & $20.1$ [$13.9$--$26.6$]          & $1.9$ [$0.0$--$4.4$] \\
Muse-Spark-1.1  & $52.6$ [$49.0$--$56.3$]          & $7.7$ [$4.9$--$11.0$]          & $\mathbf{21.5}$ [$15.3$--$28.1$] & $1.9$ [$0.0$--$4.4$] \\
GPT-5.6-Sol     & $51.5$ [$47.6$--$55.4$]          & $8.4$ [$5.2$--$12.1$]          & $17.9$ [$12.2$--$24.2$]          & $\mathbf{2.6}$ [$0.6$--$5.2$] \\
Claude-Opus-4.8 & $48.0$ [$44.4$--$51.6$]          & $4.1$ [$2.1$--$6.6$]           & $13.7$ [$8.6$--$19.5$]           & $0.7$ [$0.0$--$2.0$] \\
GLM-5.2         & $42.7$ [$39.1$--$46.5$]          & $4.0$ [$2.3$--$5.9$]           & $14.5$ [$9.4$--$20.1$]           & $0.0$ [$0.0$--$0.0$] \\
Grok-4.5        & $40.8$ [$37.4$--$44.1$]          & $4.1$ [$2.3$--$6.0$]           & $13.8$ [$8.8$--$19.4$]           & $0.0$ [$0.0$--$0.0$] \\
Kimi-K2.7-Code  & $37.0$ [$33.7$--$40.3$]          & $2.9$ [$1.4$--$4.7$]           & $11.2$ [$6.9$--$16.2$]           & $0.0$ [$0.0$--$0.0$] \\
Gemini-3.1-Pro  & $32.4$ [$29.2$--$35.7$]          & $2.6$ [$1.2$--$4.3$]           & $9.4$ [$5.0$--$13.8$]            & $0.0$ [$0.0$--$0.0$] \\
Qwen3.5-397B    & $24.4$ [$21.5$--$27.3$]          & $0.2$ [$0.0$--$0.6$]           & $1.9$ [$0.0$--$4.5$]             & $0.0$ [$0.0$--$0.0$] \\
\bottomrule
\end{tabular}%
}
\end{table}

\begin{table}[t]
\centering
\scriptsize
\caption{Performance of models on the APEX\textendash Accounting leaderboard ($n=\NTasks{}$) compared with the public dev set ($n=\DevSetTasks{}$) using Mean Criteria@\KSubsample{}.}
\label{tab:leaderboard_devset_shift}
\setlength{\tabcolsep}{4pt}
\resizebox{\columnwidth}{!}{%
\begin{tabular}{l|ccc}
\toprule
\textbf{Model}
& \makecell{\textbf{Leaderboard} \\ \textbf{Mean Criteria@\KSubsample{}}}
& \makecell{\textbf{Dev Set} \\ \textbf{Mean Criteria@\KSubsample{}}}
& \makecell{\textbf{Score} \\ \textbf{Diff}} \\
\midrule
\TopModel{}        & {\boldmath{\TopScore{}}} & $\mathbf{67.9\%}$ & $+11.5$ \\
Muse-Spark-1.1     & $52.6\%$               & $52.4\%$          & $-0.2$ \\
\ThirdModel{}      & \ThirdScore{}        & $62.3\%$          & $+10.8$ \\
\SecondModel{}     & \SecondScore{}       & $61.8\%$          & $+13.8$ \\
GLM-5.2            & $42.7\%$               & $44.1\%$          & $+1.4$ \\
Grok-4.5           & $40.8\%$               & $51.1\%$          & $+10.3$ \\
Kimi-K2.7-Code     & $37.0\%$               & $38.0\%$          & $+1.0$ \\
Gemini-3.1-Pro     & $32.4\%$               & $37.0\%$          & $+4.6$ \\
Qwen3.5-397B       & $24.4\%$               & $27.6\%$          & $+3.2$ \\
\bottomrule
\end{tabular}%
}
\end{table}

\section{Cost Ablation: Dollar-to-Token Conversion}
\label{app:cost_conversion}
This appendix section documents how the dollar budgets in the cost ablation are enforced, so the experiment can be reproduced exactly. \Cref{sec:token_ablation} enforces each per-task dollar budget (\DollarBudgetOne{}, \DollarBudgetTwo{}, \DollarBudgetThree{}, \DollarBudgetFour{}) as a token budget, converting dollars to tokens using each model's own prompt and completion token prices at a $90$:$10$ input-to-output token split, using the higher cost tier listed by model
providers (prices can vary by token count). For a model with prompt price $p$ and completion price $c$ (both in dollars per one million tokens), the token allowance for a dollar budget $B$ is
\[
  \text{tokens}(B) = \frac{B}{\left(0.9\,p + 0.1\,c\right) / 1{,}000{,}000}.
\]
\Cref{tab:cost_conversion} reports the prompt and completion price for each model in the five-model ablation roster, along with the resulting token allowance at each dollar budget, the exact token counts the \LoopHarness{} enforces directly.
\\

\begin{table}[t]
\centering
\scriptsize
\setlength{\tabcolsep}{3pt}
\caption{Prompt and completion token prices and the resulting per-task token allowance at each dollar budget, assuming a $90$:$10$ input-to-output token split, five-model ablation roster (rounded to the nearest ten).}
\label{tab:cost_conversion}
\resizebox{\columnwidth}{!}{%
\begin{tabular}{lcccccc}
\toprule
\textbf{Model} & \makecell{\textbf{Prompt Cost}\\\textbf{(\$/1M)}} & \makecell{\textbf{Completion Cost}\\\textbf{(\$/1M)}} & \textbf{\DollarBudgetOne{}} & \textbf{\DollarBudgetTwo{}} & \textbf{\DollarBudgetThree{}} & \textbf{\DollarBudgetFour{}} \\
\midrule
\TopModel{}      & \$10.00 & \$50.00 & $71{,}420$  & $357{,}140$    & $714{,}280$     & $3{,}571{,}420$  \\
\ThirdModel{}    & \$10.00 & \$45.00 & $74{,}070$  & $370{,}370$    & $740{,}740$     & $3{,}703{,}700$  \\
Gemini-3.1-Pro   & \$4.00  & \$18.00 & $185{,}180$ & $925{,}920$    & $1{,}851{,}850$ & $9{,}259{,}250$  \\
Grok-4.5         & \$4.00  & \$12.00 & $208{,}330$ & $1{,}041{,}660$ & $2{,}083{,}330$ & $10{,}416{,}660$ \\
Muse-Spark-1.1   & \$1.25  & \$4.25  & $645{,}160$ & $3{,}225{,}800$ & $6{,}451{,}610$ & $32{,}258{,}060$ \\
\bottomrule
\end{tabular}%
}
\end{table}

\Cref{fig:cost_prompt_example} shows an example of how this allowance is surfaced to the model: the converted token budget is stated verbatim in the system prompt's Token Budget section, so the model can pace its own tool calls and reasoning against a concrete, known limit rather than an opaque dollar figure. The remaining token budget is also shared with the model at each step. 

\begin{figure}[t]
\centering
\caption{System-prompt excerpt disclosing the converted token budget to the model, shown for Grok-4.5 at the \DollarBudgetTwo{} allowance from \Cref{tab:cost_conversion}. Models are informed of their token budget the same way they are informed of the per-run step budget (\Cref{sec:token_ablation}).}

\begin{tcolorbox}[colback=gray!5, colframe=black!60, width=\columnwidth,
  title={Example System Prompt Excerpt (Grok-4.5, \DollarBudgetTwo{} Budget)}, fonttitle=\bfseries]
\scriptsize\ttfamily
You are an AI assistant that completes accounting tasks by reasoning and using tools.

\vspace{0.5em}\textbf{\#\# Token Budget}\\
You have a total of 1,041,660 tokens to use throughout the run. If you run out of tokens, submit what you have for partial credit.

\vspace{0.5em}\textbf{\#\# Think Before Acting}\\
Before making tool calls, briefly explain your reasoning in 1--3 sentences: what you learned from the previous step, and what you are doing next and why. Be concise but show your thinking.

\vspace{0.5em}\normalfont\itshape[\ldots{} standard task-file, tool-use, workflow, and output-format instructions omitted \ldots]
\end{tcolorbox}
\label{fig:cost_prompt_example}
\end{figure}

\section{Within-Cap-Spend Score Analysis}
\label{app:spend_score_slope}

This appendix supports the cost ablation in \Cref{sec:token_ablation}. The paired per-task budget gains are reported in \Cref{tab:paired_deltas}. \Cref{tab:panel_b_9010} reports the within-cap levels slope $b$ (change in Mean Criteria@\KSubsample{} per dollar of realized spend) for each model$\times$budget cell. Because we set the budget cap but not how much of it a model spends on any given task, these slopes describe associations rather than causal effects; the most plausible driver is task difficulty (hard tasks draw more tokens and still score lower), which our design can note but not confirm.

\begin{table*}[t]
\centering
\small
\setlength{\tabcolsep}{8pt}
\renewcommand{\arraystretch}{1.15}
\caption{Paired per-task quality gains from raising the dollar budget (Mean Criteria@\KSubsample{}, percentage points), with $95\%$ bootstrap CIs. $p$-values are Benjamini--Hochberg corrected across the $20$ model$\times$contrast cells; gains significant at the $5\%$ false-discovery-rate level are shown in \textbf{bold}.}
\label{tab:paired_deltas}
\resizebox{\textwidth}{!}{%
\begin{tabular}{lcccc}
\toprule
\textbf{Model} & \textbf{$\Delta$(\$5--\$1)} & \textbf{$\Delta$(\$10--\$5)} & \textbf{$\Delta$(\$50--\$10)} & \textbf{$\Delta$(\$50--\$1)} \\
\midrule
\TopModel{}    & $\mathbf{+30.4}$ [$+26.8$, $+34.2$] & $\mathbf{+6.3}$ [$+3.7$, $+8.9$]  & $\mathbf{+6.7}$ [$+4.1$, $+9.4$]  & $\mathbf{+43.4}$ [$+39.1$, $+47.7$] \\
\ThirdModel{}  & $\mathbf{+11.2}$ [$+8.3$, $+14.3$]  & $-0.2$ [$-2.5$, $+2.1$]       & $+1.3$ [$-0.8$, $+3.5$]       & $\mathbf{+12.3}$ [$+9.2$, $+15.5$] \\
Gemini-3.1-Pro & $\mathbf{+5.4}$ [$+3.1$, $+7.8$]   & $-0.5$ [$-2.7$, $+1.8$]       & $\mathbf{+4.2}$ [$+2.1$, $+6.3$]  & $\mathbf{+9.1}$ [$+6.5$, $+11.8$] \\
Muse-Spark-1.1 & $\mathbf{+4.1}$ [$+2.0$, $+6.3$]    & $+0.5$ [$-1.5$, $+2.5$]       & $+0.0$ [$-2.2$, $+2.2$]       & $\mathbf{+4.7}$ [$+2.3$, $+7.0$] \\
Grok-4.5       & $\mathbf{+3.5}$ [$+0.7$, $+6.4$]         & $-1.8$ [$-4.8$, $+1.2$]       & $+1.1$ [$-1.9$, $+4.2$]       & $+2.8$ [$-0.2$, $+5.9$] \\
\bottomrule
\end{tabular}%
}
\end{table*}

\begin{table*}[t]
\centering
\small
\setlength{\tabcolsep}{8pt}
\renewcommand{\arraystretch}{1.15}
\caption{Within-cap levels slope $b$ (Mean Criteria@\KSubsample{} per USD) for each model$\times$budget cell, $90$:$10$ input:output split, with $95\%$ bootstrap CIs. $p$-values are Benjamini--Hochberg corrected across the $20$ cells; slopes significant at the $5\%$ false-discovery-rate level are shown in \textbf{bold}. Slopes are descriptive associations, not causal effects.}
\label{tab:panel_b_9010}
\resizebox{\textwidth}{!}{%
\begin{tabular}{lccccc}
\toprule
\textbf{Budget} & \textbf{\TopModel{}} & \textbf{\ThirdModel{}} & \textbf{Gemini-3.1-Pro} & \textbf{Muse-Spark-1.1} & \textbf{Grok-4.5} \\
\midrule
\$1  & $0.065$ [$-0.016$, $0.231$]        & $0.015$ [$-0.027$, $0.060$]          & $-0.069$ [$-0.247$, $0.051$]      & $-0.136$ [$-0.300$, $0.013$]     & $-0.069$ [$-0.202$, $0.080$] \\
\$5  & $-0.101$ [$-0.167$, $-0.020$]      & $\mathbf{-0.031}$ [$-0.052$, $-0.010$]   & $-0.033$ [$-0.059$, $-0.007$]     & $-0.017$ [$-0.044$, $0.009$]     & $0.003$ [$-0.023$, $0.031$] \\
\$10 & $\mathbf{-0.060}$ [$-0.084$, $-0.032$] & $\mathbf{-0.020}$ [$-0.033$, $-0.007$]  & $-0.012$ [$-0.025$, $0.000$]     & $-0.008$ [$-0.023$, $0.006$]     & $0.003$ [$-0.009$, $0.016$] \\
\$50 & $-0.002$ [$-0.005$, $0.001$]       & $-0.001$ [$-0.004$, $0.001$]         & $-0.001$ [$-0.003$, $0.000$]      & $-0.004$ [$-0.015$, $0.003$]     & $-0.003$ [$-0.008$, $0.002$] \\
\bottomrule
\end{tabular}%
}
\end{table*}

\section{Harness Comparison Details}
\label{app:harness_details}
This comparison tests whether a purpose-built accounting harness affects model scores once task difficulty is controlled. \Cref{tab:harness_comparison} reports Mean Criteria@\KSubsample{} and Pass@1 for every model under the \LoopHarness{} and the \RampHarness{}, with task-bootstrap $95\%$ CIs over the shared $\NTasks{}$-task set ($154$--$160$ paired tasks per model after excluding incomplete runs); \ThirdModel{} has no \RampHarness{} run. Of the eight models with both harnesses, only Grok-4.5's Mean Criteria@\KSubsample{} gain survives Benjamini--Hochberg correction across the $16$ paired contrasts ($+7.5$\,pp, CI $[5.0, 10.2]$, $q < 10^{-6}$; $61\%$ of its tasks improve, $25\%$ worsen). Three other shifts exclude zero nominally but do not survive BH correction: Muse-Spark-1.1's Mean Criteria@\KSubsample{} decline ($-3.1$\,pp, CI $[-6.2, -0.2]$), \SecondModel{}'s gain ($+2.2$\,pp, CI $[0.1, 4.2]$), and Muse-Spark-1.1's Pass@1 gain ($+2.5$\,pp, CI $[0.2, 5.0]$); all shifts are paired-task deltas over shared tasks. 

\begin{table}[t]
\centering
\scriptsize
\caption{Mean Criteria@\KSubsample{} and Pass@1 (\%) under the \LoopHarness{} vs.\ the \RampHarness{} ($k=\KSubsample{}$, $n=\NTasks{}$). Brackets are task-bootstrap $95\%$ CIs; GPT-5.6-Sol has no \RampHarness{} run and is omitted. The one Benjamini--Hochberg-surviving gain (Grok-4.5, Mean Criteria@\KSubsample{}) is in \textbf{bold}.}
\label{tab:harness_comparison}
\setlength{\tabcolsep}{2.5pt}
\renewcommand{\arraystretch}{1.05}
\resizebox{\columnwidth}{!}{%
\begin{tabular}{lcccc}
\toprule
 & \multicolumn{2}{c}{\textbf{Mean Criteria@\KSubsample{}}} & \multicolumn{2}{c}{\textbf{Pass@1}} \\
\cmidrule(lr){2-3}\cmidrule(lr){4-5}
\textbf{Model} & \textbf{Loop} & \textbf{Ramp} & \textbf{Loop} & \textbf{Ramp} \\
\midrule
Claude-Fable-5            & $56.4$ [$52.7$--$60.2$] & $55.6$ [$51.7$--$59.4$] & $9.5$ [$6.0$--$13.4$] & $10.0$ [$6.0$--$14.4$] \\
Muse-Spark-1.1     & $52.6$ [$48.9$--$56.2$] & $49.8$ [$45.5$--$54.0$] & $7.7$ [$4.9$--$11.0$] & $10.2$ [$6.5$--$14.4$] \\
Claude-Opus-4.8    & $48.0$ [$44.5$--$51.5$] & $50.4$ [$46.8$--$53.9$] & $4.1$ [$2.1$--$6.6$]  & $5.8$ [$3.1$--$9.0$] \\
GLM-5.2            & $42.7$ [$38.9$--$46.4$] & $43.2$ [$39.7$--$46.7$] & $4.0$ [$2.3$--$5.9$]  & $4.2$ [$2.1$--$6.7$] \\
Grok-4.5           & $40.8$ [$37.5$--$44.2$] & $\mathbf{48.3}$ [$44.7$--$52.0$] & $4.1$ [$2.3$--$6.0$]  & $5.6$ [$2.9$--$8.8$] \\
Kimi-K2.7-Code     & $37.0$ [$33.7$--$40.3$] & $37.5$ [$34.2$--$40.8$] & $2.9$ [$1.4$--$4.7$]  & $2.5$ [$0.8$--$4.6$] \\
Gemini-3.1-Pro     & $32.4$ [$29.1$--$35.7$] & $33.7$ [$30.4$--$37.2$] & $2.6$ [$1.2$--$4.3$]  & $3.5$ [$1.5$--$6.2$] \\
Qwen3.5-397B       & $24.4$ [$21.5$--$27.3$] & $25.3$ [$22.4$--$28.4$] & $0.2$ [$0.0$--$0.6$]  & $0.2$ [$0.0$--$0.6$] \\
\bottomrule
\end{tabular}%
}
\end{table}

\subsection{Subagent Delegation}
\label{app:subagent}
\begin{figure}[t]
\centering
\caption{Mean subagents spawned per task by model under the \RampHarness{} (task-bootstrap $95\%$ CIs).}
\includegraphics[width=\columnwidth]{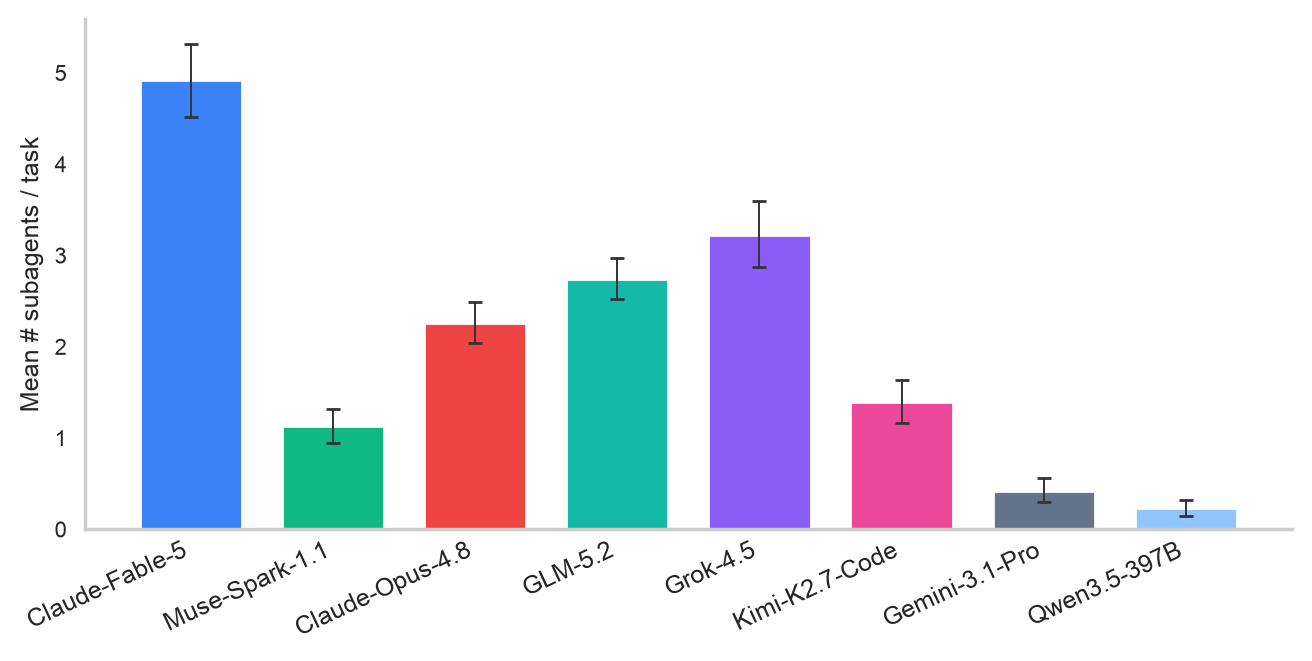}
\label{fig:subagents_per_task}
\end{figure}

\Cref{fig:subagents_per_task} shows how heavily each model delegates work: \TopModel{} spawns subagents in $89\%$ of its trajectories ($4.9$ per task, CI $[4.5, 5.3]$); GLM-5.2, \SecondModel{}, and Grok-4.5 follow at $69$--$89\%$ ($2.3$--$3.2$ per task); Kimi-K2.7-Code and Muse-Spark-1.1 delegate in $37$--$40\%$ ($1.1$--$1.4$ per task); and Gemini-3.1-Pro and Qwen3.5-397B delegate rarely ($12$--$13\%$ of trajectories, $0.2$--$0.4$ per task). \\

Spawn intensity does not line up with harness-level gains: \TopModel{} and GLM-5.2 delegate most heavily yet shift by less than a point under the \RampHarness{} ($-0.8$ and $+0.5$\,pp), while Grok-4.5, the only model with a Benjamini--Hochberg-surviving gain (\Cref{tab:harness_comparison}), delegates in $69\%$ of its trajectories. Heavy delegation is therefore attributed to model preference rather than performance gains.

\section{Failure Modes of the Best-Performing Models}
\label{app:failure_by_top_models}
\begin{figure*}[p]
\centering
\caption{Failure-mode breakdown for the three best-performing models ($n=24$/$22$/$28$ annotated failures). Inner ring: L1 failure categories; outer ring: L2 subclasses.}
\begin{subfigure}{0.48\textwidth}
\centering
\caption{Claude-Fable-5}
\includegraphics[width=\textwidth]{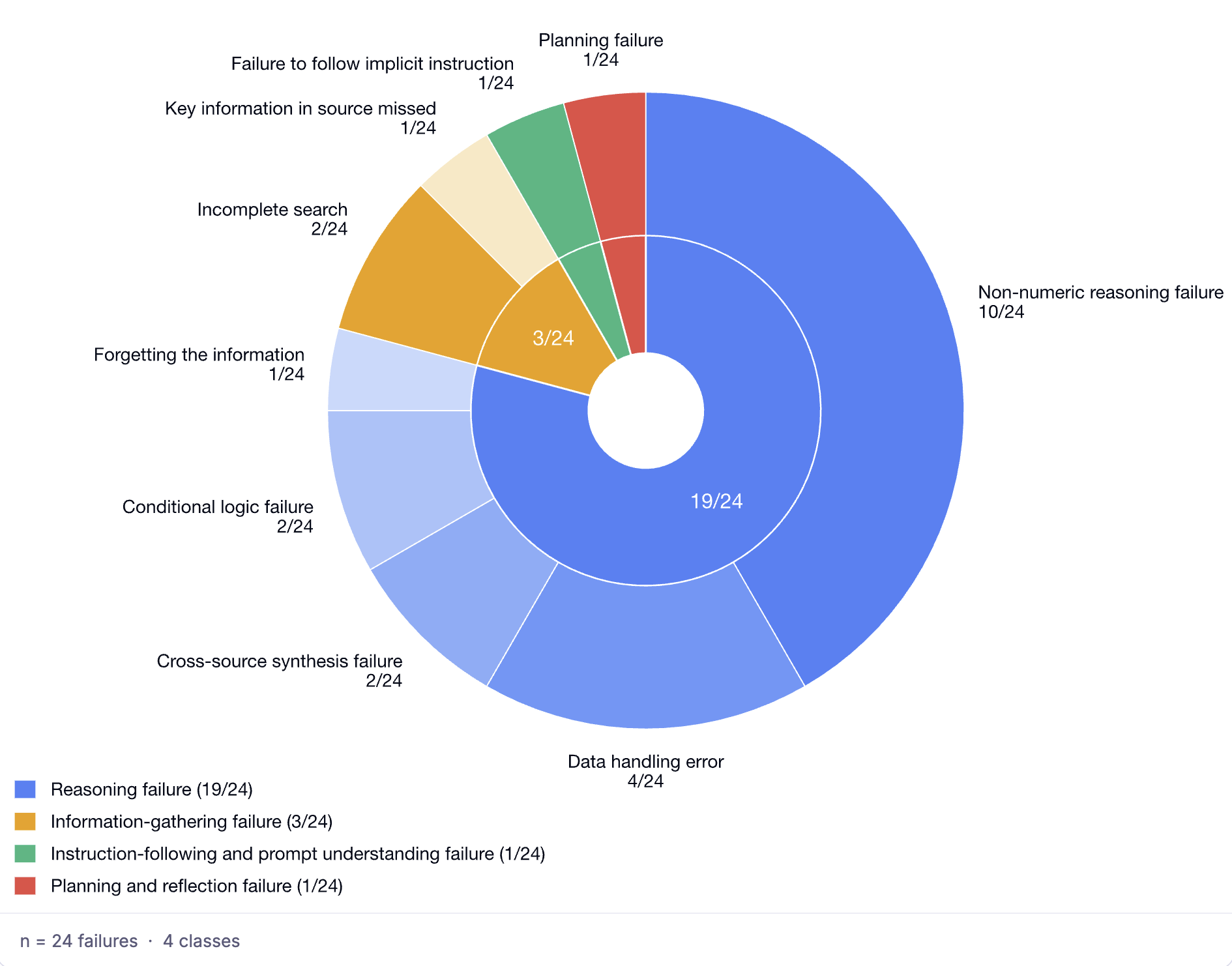}
\label{fig:sunburst_fable}
\end{subfigure}
\hfill
\begin{subfigure}{0.48\textwidth}
\centering
\caption{Muse-Spark-1.1}
\includegraphics[width=\textwidth]{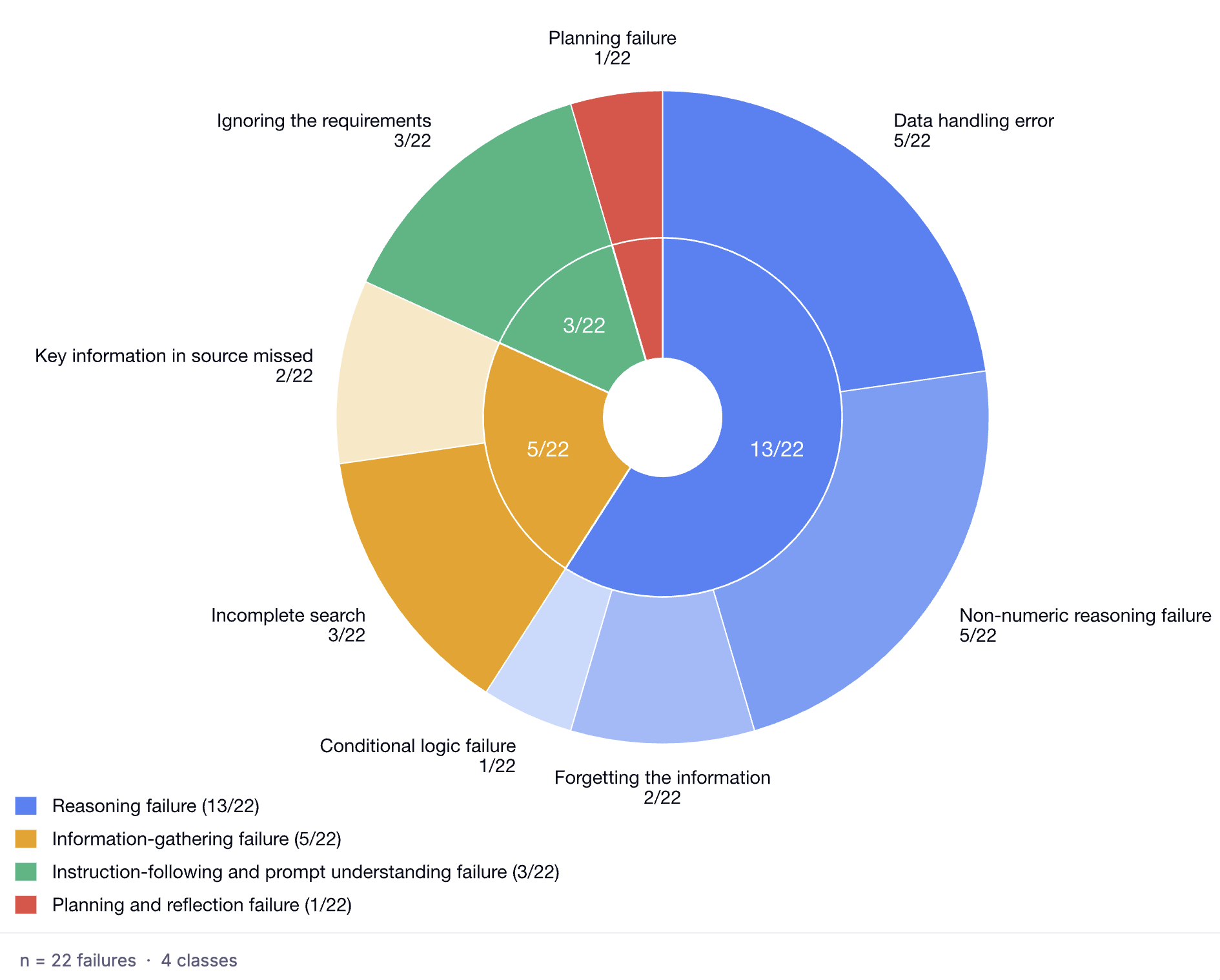}
\label{fig:sunburst_muse}
\end{subfigure}

\vspace{0.6em}

\begin{subfigure}{0.48\textwidth}
\centering
\caption{GPT-5.6-Sol}
\includegraphics[width=\textwidth]{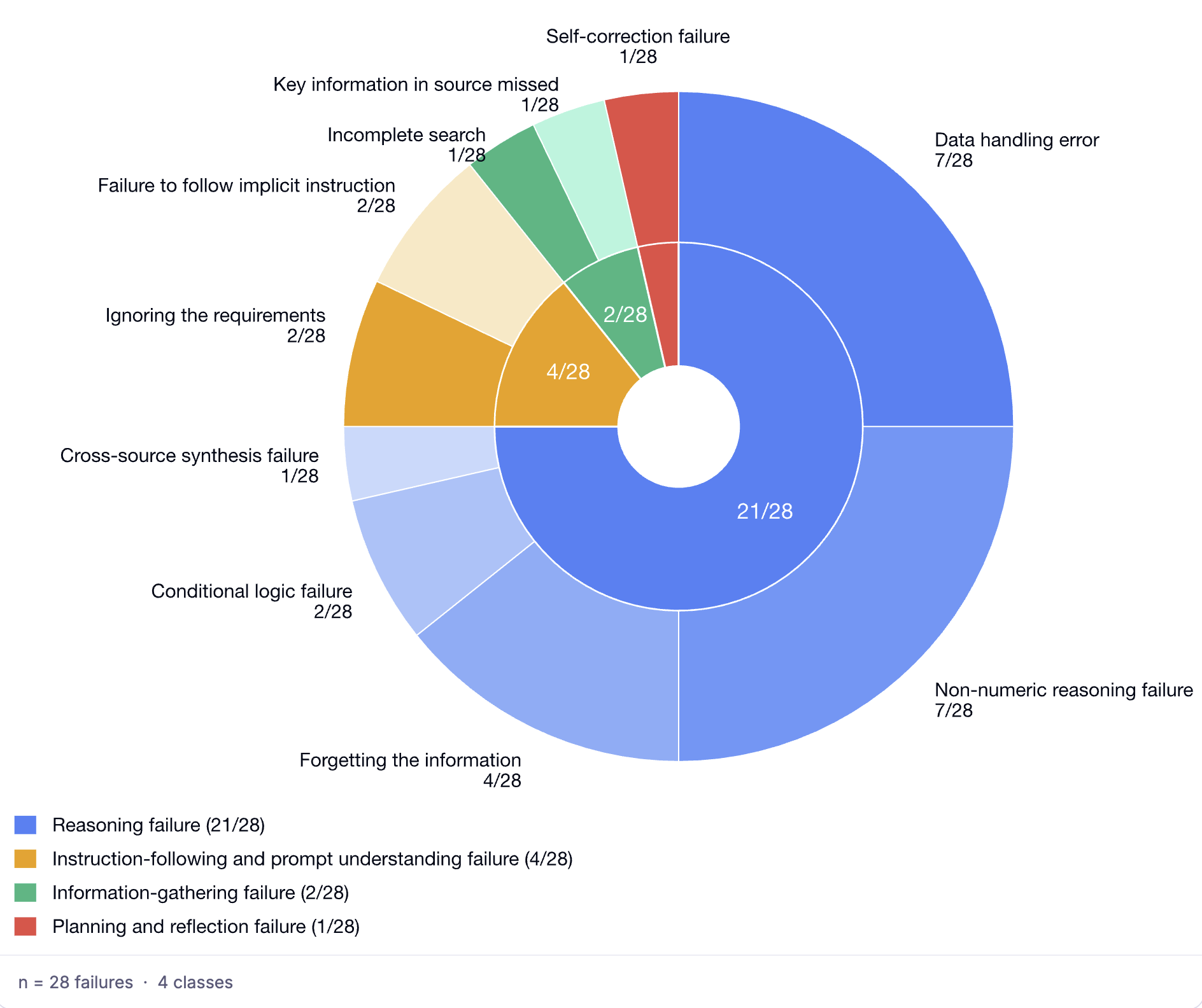}
\label{fig:sunburst_gpt}
\end{subfigure}

\label{fig:failure_sunbursts_top3}
\end{figure*}

\Cref{sec:failure_modes_top_models} compares the top models at the L1 level and within reasoning failures; \Cref{fig:failure_sunbursts_top3} gives the complete picture, breaking every annotated failure down to its L2 subclass per model. Inner rings show L1 failure categories; outer rings show the L2 subclasses within each. Beyond the shared concentration in reasoning, the sunbursts show where the profiles diverge: Claude-Fable-5 concentrates hardest in reasoning ($79\%$ of its failures, over half of them non-numeric reasoning), GPT-5.6-Sol carries the broadest instruction-following tail (ignoring requirements and implicit-instruction failures), and Muse-Spark-1.1 is the only model whose information-gathering share exceeds a fifth.

\vspace{0.4em}

\subsection{L2 Subclass Counts}
\label{app:failure_l2_counts}
\Cref{tab:failure_l2_counts} reports the exact per-model counts behind \Cref{fig:failure_sunbursts_top3}. Non-numeric reasoning failures and data handling errors are the two largest subclasses for every model, and no annotated failure involves tool use.

\begin{table*}[t]
\centering
\small
\renewcommand{\arraystretch}{1.15}
\caption{Annotated failure counts by L2 subclass for the three best-performing models, grouped by L1 category ($n=24$/$28$/$22$; $74$ failures total). The four L1 categories shown are the only ones observed; no annotated failure falls under tool use or any other taxonomy branch.}
\label{tab:failure_l2_counts}
\begin{tabular}{@{}lcccc@{}}
\toprule
\textbf{L2 subclass} & \textbf{Claude-Fable-5} & \textbf{GPT-5.6-Sol} & \textbf{Muse-Spark-1.1} & \textbf{Total} \\
\midrule
\rowcolor{gray!15}
\multicolumn{5}{@{}l}{\textit{Reasoning failure}} \\
\quad Non-numeric reasoning failure & $10$ & $7$ & $5$ & $22$ \\
\quad Data handling error & $4$ & $7$ & $5$ & $16$ \\
\quad Forgetting the information & $1$ & $4$ & $2$ & $7$ \\
\quad Conditional logic failure & $2$ & $2$ & $1$ & $5$ \\
\quad Cross-source synthesis failure & $2$ & $1$ & $0$ & $3$ \\
\midrule
\rowcolor{gray!15}
\multicolumn{5}{@{}l}{\textit{Information-gathering failure}} \\
\quad Incomplete search & $2$ & $1$ & $3$ & $6$ \\
\quad Key information in source missed & $1$ & $1$ & $2$ & $4$ \\
\midrule
\rowcolor{gray!15}
\multicolumn{5}{@{}l}{\textit{Instruction-following and prompt understanding failure}} \\
\quad Ignoring the requirements & $0$ & $2$ & $3$ & $5$ \\
\quad Failure to follow implicit instruction & $1$ & $2$ & $0$ & $3$ \\
\midrule
\rowcolor{gray!15}
\multicolumn{5}{@{}l}{\textit{Planning and reflection failure}} \\
\quad Planning failure & $1$ & $0$ & $1$ & $2$ \\
\quad Self-correction failure & $0$ & $1$ & $0$ & $1$ \\
\midrule
\textbf{Total} & $24$ & $28$ & $22$ & $74$ \\
\bottomrule
\end{tabular}
\end{table*}

\section{Failure Taxonomy}
\label{app:taxonomy}
\Cref{sec:failure_modes} introduces the taxonomy at a high level. This appendix section lists the taxonomy at the L1/L2 level. $7$ L1 categories and $39$ L2 subclasses are relevant to accounting work; finer L3 sub-labels are omitted for brevity.

\subsection*{1. Information-gathering failure}
\textit{The agent fails to find the correct information, or all of the information, required to successfully complete the request.}
\begin{itemize}[nosep, leftmargin=1.2em]
\item \textbf{No information gathering.} The agent does not attempt to search or retrieve any files at any point in the task, relying on parametric knowledge to complete the request.
\item \textbf{Incomplete search.} The agent searches for or retrieves only some of the required information: it never finds the correct source or finds a similar but poorer-quality source.
\item \textbf{Unused search result.} The agent finds the correct source in its search results but fails to open or use it.
\item \textbf{Key information in source missed.} The agent retrieves the correct file(s) but does not find the relevant or correct information within them.
\item \textbf{Source reconstruction error.} The agent reconstructs a value from an original source (usually raw inputs) and introduces an inconsistency.
\item \textbf{Grounding failure -- incorrect source attribution.} The agent attributes information to a source that does not actually support or contain it.
\end{itemize}

\subsection*{2. Instruction-following and prompt understanding failure}
\textit{The agent fails to understand or apply some or all of the instructions and constraints in the prompt.}
\begin{itemize}[nosep, leftmargin=1.2em]
\item \textbf{Ignoring the requirements.} The agent ignores one or more of the prompt's requirements, constraints, or dimensions, such as a time period, business segment, or entity.
\item \textbf{Failure to follow implicit instruction.} The agent fails to follow, ignores, or misinterprets an implicit instruction in the prompt.
\item \textbf{Priority error.} The agent applies an incorrect priority order among multiple satisfiable instructions that compete for attention, resources, or scope.
\item \textbf{Ignores a negation.} The agent ignores an exclusionary or negating term such as ``not,'' ``except,'' or ``excluding'' and acts as though it were absent.
\item \textbf{Non-text interpretation failure.} The agent misreads or fails to extract information from non-text content in the prompt, such as tables, graphs, images, or images of text (e.g., scanned documents or screenshots).
\end{itemize}

\subsection*{3. Reasoning failure}
\textit{The agent has the correct information available but applies faulty logic, inference, or calculation to it.}
\begin{itemize}[nosep, leftmargin=1.2em]
\item \textbf{Numeric reasoning failure.} The agent makes an incorrect calculation or applies faulty quantitative logic to numeric data; the underlying values may be correct and complete, but the arithmetic or quantitative method applied to them is not.
\item \textbf{Data handling error.} The agent correctly understands the source data's structure and content but selects an incorrect method to filter, join, aggregate, or otherwise process it.
\item \textbf{Non-numeric reasoning failure.} The agent applies faulty logic or inference to non-numeric information, reaching an incorrect conclusion even though the available information is sufficient.
\item \textbf{Cross-source synthesis failure.} The agent fails to correctly and completely link or reconcile information across sources into a single accurate value or perspective.
\item \textbf{Exception blindness.} The agent applies a general rule, default approach, or commonly used value without recognizing or checking for a documented exception, carve-out, or methodology.
\item \textbf{Conditional logic failure.} The agent identifies an if/then rule but misapplies or ignores it: acting when it should not, failing to act when it should, or applying the wrong outcome to a correctly identified condition.
\item \textbf{Failure to request clarification.} When the prompt contradicts available evidence, rests on an incorrect premise, requests something infeasible, or includes a mistake by the user, the agent fails to seek clarification before attempting the task.
\item \textbf{Failure to proactively correct.} The agent fails to resolve confusion, ambiguity, or mistakes in the prompt that could be addressed without input from the user.
\item \textbf{Forgetting the information.} The agent fails to carry its own correct prior reasoning, calculations, or intermediate conclusions into subsequent steps, contradicting, abandoning, or replacing them without new evidence or reconciling the discrepancy.
\item \textbf{Disambiguation failure.} The agent fails to disambiguate between similar technical terms with different meanings in context, selecting the wrong meaning or treating the terms as interchangeable.
\item \textbf{Missed dependency.} The agent fails to identify dependencies between steps, or identifies a dependency but does not sequence or handle it correctly.
\end{itemize}                               

\subsection*{4. Tool use failure}
\textit{The agent fails to use any tools, identifies the correct tool but does not use it, uses an incorrect tool, or uses the correct tool incorrectly.}
\begin{itemize}[nosep, leftmargin=1.2em]
\item \textbf{No tool used.} The agent does not identify a tool to use to complete an action, relying solely on parametric knowledge.
\item \textbf{Tool not used.} The agent identifies the correct tool but does not actually call, load, or execute it.
\item \textbf{Incorrect tool used.} The agent selects a tool that cannot perform the function the task requires.
\item \textbf{Tool used incorrectly.} The agent selects the correct tool but operates it in a way that produces a faulty result or no result.
\item \textbf{Tool output misinterpretation.} The agent fails to correctly understand the results, errors, or status returned by a tool.
\end{itemize}

\subsection*{5. Planning and reflection failure}
\textit{The agent fails to appropriately plan, monitor, or adjust its process while completing the request.}
\begin{itemize}[nosep, leftmargin=1.2em]
\item \textbf{Planning failure.} The agent fails to plan, or to course-correct, a viable course of action for completing the request.
\item \textbf{Failure to complete task.} The agent enters a doom loop, repeating the same or functionally similar steps, and cannot exit the loop to complete the task.
\item \textbf{Verification failure.} The agent does not verify its final output before submission, or verifies it only partially; this applies at the point of the final response. It does not apply at intermediate steps.
\item \textbf{Self-correction failure.} The agent makes an error while completing the request but fails to identify or correct it.
\item \textbf{Failure to retry.} The agent does not retry or opt for an appropriate, available fallback when a step fails.
\end{itemize}

\subsection*{6. Communication and presentation failure}
\textit{The agent makes a style, grammar, formatting, or other communication error.}
\begin{itemize}[nosep, leftmargin=1.2em]
\item \textbf{Final output mistake.} The agent correctly completes all reasoning and analysis but misstates the results in its output.
\item \textbf{Tone failure.} The agent fails to adapt its tone or formality to the context, audience, or a specific user request.
\item \textbf{Language failure.} The agent makes a typographical, grammar, or sentence-construction error in its final output.
\item \textbf{Excessively verbose response.} The agent's output contains unnecessary length or repetition, egregious enough to make the output considerably harder to read.
\item \textbf{Overly concise response.} The agent's output omits key information it had already gathered or derived, leaving the response too brief to be useful or complete.
\item \textbf{Overconfidence.} The agent presents its response as a certainty when the available evidence does not support that level of confidence.
\item \textbf{Output format failure.} The agent's output is not structured or presented in the format(s) requested in the prompt.
\end{itemize}

\subsection*{7. Other valid failure}
\textit{The agent fails in a way that is not captured by any other failure label.}

\section{LM Usage}
We used LMs to assist with drafting and refinement of this paper.

\end{document}